\documentclass{article}

\usepackage{arxiv}

\usepackage[utf8]{inputenc} 
\usepackage[T1]{fontenc}    
\usepackage{hyperref}       
\usepackage{url}            
\usepackage{booktabs}       
\usepackage{amsfonts}       
\usepackage{nicefrac}       
\usepackage{microtype}      

\usepackage{graphicx}
\usepackage{natbib}

\usepackage{subcaption} 
\usepackage{amsmath} 
\usepackage{amsthm} 
\usepackage{array}

\usepackage{cleveref}       

\newtheorem*{probabilistic_actual_causation}{Probabilistic Actual Causation}

\newcolumntype{C}[1]{>{\centering\arraybackslash}p{#1}}

\title{Robot Pouring: Identifying  Causes of Spillage and Selecting Alternative Action Parameters Using Probabilistic Actual Causation}

\newif\ifuniqueAffiliation
\uniqueAffiliationtrue

\ifuniqueAffiliation 
\author{ Jaime Maldonado \\
	Cognitive Neuroinformatics\\
	University of Bremen\\
	Bremen, Germany \\
	\texttt{jmaldonado@uni-bremen.de} \\
	\And
	Jonas Krumme \\
	Cognitive Neuroinformatics\\
	University of Bremen\\
	Bremen, Germany \\
	\And
	Christoph Zetzsche \\
	Cognitive Neuroinformatics\\
	University of Bremen\\
	Bremen, Germany \\
	\And
	Vanessa Didelez \\
	Department of Biometry and Data Management\\
	Leibniz Institute for Prevention Research and Epidemiology - BIPS\\
	Bremen, Germany \\
	\And
	Kerstin Schill \\
	Cognitive Neuroinformatics\\
	University of Bremen\\
	Bremen, Germany \\
}

\renewcommand{\headeright}{ }
\renewcommand{\undertitle}{ }
\renewcommand{\shorttitle}{Robot Pouring - Probabilistic Actual Causation}

\hypersetup{
	pdftitle={Robot Pouring - Probabilistic Actual Causation},
	pdfsubject={},
	pdfauthor={},
	pdfkeywords={robot pouring, causality, probabilistic actual causation, causal discovery, action-guiding explanations},
}

\begin{document}
	\date{}
	\maketitle
	
	\begin{abstract}
	In everyday life, we perform tasks (e.g., cooking or cleaning) that involve a large variety of objects and goals. When confronted with an unexpected or unwanted outcome, we take corrective actions and try again until achieving the desired result. The reasoning performed to identify a cause of the observed outcome and to select an appropriate corrective action is a crucial aspect of human reasoning for successful task execution. Central to this reasoning is the assumption that a factor is responsible for producing the observed outcome. In this paper, we investigate the use of probabilistic actual causation to determine whether a factor is the cause of an observed undesired outcome. Furthermore, we show how the actual causation probabilities can be used to find alternative actions to change the outcome. We apply the probabilistic actual causation analysis to a robot pouring task. When spillage occurs, the analysis indicates whether a task parameter is the cause and how it should be changed to avoid spillage. The analysis requires a causal graph of the task and the corresponding conditional probability distributions. To fulfill these requirements, we perform a complete causal modeling procedure (i.e., task analysis, definition of variables, determination of the causal graph structure, and estimation of conditional probability distributions) using data from a realistic simulation of the robot pouring task, covering a large combinatorial space of task parameters. Based on the results, we discuss the implications of the variables' representation and how the alternative actions suggested by the actual causation analysis would compare to the alternative solutions proposed by a human observer. The practical use of the analysis of probabilistic actual causation to select alternative action parameters is demonstrated. 
	\end{abstract}

	
	\section{Introduction}
	Pouring the content of a source container into a target container requires planning, perception, and action capabilities. Humans excel at these capabilities and can skillfully pour any material into containers of arbitrary shapes and dimensions. If spillage occurs, we can take corrective actions (e.g., selecting a target container with an appropriate capacity) and try again to pour without spilling. The ability to take corrective actions is a crucial aspect of human reasoning for successful task execution. Implementing similar reasoning capabilities in robotic systems can reduce task failures and enable a robust operation in unstructured environments.
	
	In the pouring task, possible causes for the undesired outcome of spilling the poured material include the low capacity of the target container or the diameter difference of the containers (imagine pouring from a wide glass into the narrow mouth of a bottle), among others. It is reasonable to assume that corrective actions we take in everyday life target the perceived actual cause of the undesired outcome. The extreme opposite of this behavior would consist of randomly changing action variables and observing the task outcome until finding a suitable solution. For example, if we perceive that the cause of spillage is the rim of the target container being too narrow, the corrective action would consist of selecting a target container with a wider rim. Implementing similar reasoning capabilities on robotic or automatic systems would require 1) a mechanism to identify the (not necessarily unique) actual cause of an observed outcome and 2) a principled way to determine how the causal variable needs to be changed to obtain a different outcome.  
	
	Within the context of causal analysis and modeling methods, the concept of \textit{actual cause} refers to the conditions under which a particular event is recognized to be responsible for producing an outcome~\citep{Pearl2009ActualCausation}. Definitions of actual causation have been mainly used in practical applications to generate explanations for observed outcomes (see works reviewed in Section~\ref{sec:related_work}). It has been argued that explanations obtained from the analysis of actual causation can guide the search for alternative actions aiming to change the observed outcome~\citep{Beckers2022}. The practical utility of the analysis of actual causation for the search and selection of alternative actions remains to be explored in real-life applications. 
	
	In this paper, we explore the use of actual causation analysis to select action parameters in a robot pouring task. The aim is to identify the actual cause of spillage and to determine how a task parameter should be changed to pour without spilling. We use the probabilistic actual causation definition~\citep{FentonGlynn2021} to identify the cause of spillage among the variables involved in the task. In a series of examples, we illustrate how the analysis of actual causation can be used to select alternative task parameters. Section~\ref{sec:actual_causality_framework} introduces the probabilistic actual causation framework and its potential use for action guidance. In Section~\ref{sec:how_to_use_actual_causality}, we propose a procedure to use the actual causation probabilities as a principled criterion to find alternative actions.   
	
	The robot pouring task was implemented in a simulation (described in Section~\ref{sec:robot_simulation}) using a physics engine for realistic behavior. The simulation enabled us to generate trials covering a large combinatorial space of trial parameters (fullness levels and container properties), which would have been cumbersome to achieve in a physical setup. There are two prerequisites to perform an analysis of probabilistic actual causation: 1) a causal graph of the system and 2) the conditional probability distributions necessary to compute interventional queries (i.e., ``do" operations ). The variables used to represent the pouring task as a causal graph are described in Section~\ref{sec:variables}. To obtain the graph structure, we used a causal discovery algorithm (Section~\ref{sec:discovery}). The conditional distributions used to compute the do-operations were estimated using neural networks (Section~\ref{sec:NADE}). Subsequently, the causal probability and actual causation expressions that result from the graph structure are presented in Section~\ref{sec:dag_and_formulas}.  
	
	Considering that causal probability computations cannot tell per se what caused an observed outcome as this may be specific to the given circumstances, in Section~\ref{sec:causal_effect_results}, we analyze the causal probability functions for interventions on each variable to gain insight into their influence on spillage and to emphasize their limitations for action guidance.
		Subsequently, in Sections \ref{sec:actual_causality_results} and \ref{sec:evaluation_corrective_actions} we demonstrate that the analysis based on actual causation can be used in a principled way to select alternative action parameters. Firstly, in Section \ref{sec:actual_causality_results} we present four detailed examples of applying actual causation analysis on spillage trials to select alternative action parameters. The examples illustrate how different alternative actions yield lower or higher probabilities of spillage. Secondly, in Section \ref{sec:evaluation_corrective_actions} we evaluate the likelihood of finding an alternative value and the pouring success rates obtained when running the spillage trials using alternative values. The evaluation was conducted using a test dataset acquired in simulation. Overall, Sections \ref{sec:actual_causality_results} and \ref{sec:evaluation_corrective_actions} provide empirical evidence of the practical usefulness of the actual causation approach to identify alternative parameters to prevent spillage.
	
	Finally, in Section~\ref{sec:discussion}, we discuss the implications of the variables' representation and how the alternative actions suggested by the actual causation analysis would compare to the alternative solutions proposed by a human observer.
	
	In summary, our contributions are:
		\begin{itemize}
			\item We report a complete analysis of probabilistic actual causation on a practical problem with all the necessary steps for its implementation (analysis of the task, variable definition,  causal-graph structure, and estimation of causal probabilities using neural networks).
			\item We show how to use the explanations obtained with the analysis of probabilistic actual causation for action guidance in a practical use case. By doing this, we go beyond the basic diagnostic (attribution of actual cause) toward action guidance. 
			\item We evaluate the capability of the actual causation approach  to automatically identify alternative parameters of different variables.
			\item We evaluate the pouring success rates obtained when the alternative parameters are used to prevent spillage.
		\end{itemize}

	\section{Probabilistic Actual Causation and Action-Guiding Explanations}\label{sec:actual_causality_framework}
	The concept of actual cause refers to the conditions under which a particular event is recognized to be responsible for producing an outcome in a specific scenario or context~\citep{Pearl2009ActualCausation}. The purpose of determining an actual cause is to find which past actions explain an already observed output~\citep{Beckers2022}. The definition of actual causation proposed by \citet{Halpern2005} for deterministic scenarios is one of the most prominent definitions in the literature~\citep{Borner2023}. \citet{FentonGlynn2017,FentonGlynn2021} proposed an extension of Halpern and Pearl's definition that is apt for probabilistic causal scenarios. The Fenton-Glynn's definition of actual causation is formulated in the framework of probabilistic causal models, where the causal model is a causal Bayesian network represented graphically as a directed acyclic graph (DAG) and the link between each node/variable and its direct causes is modeled probabilistically~\citep{Pearl2009DAGs}. In the DAG formalism, nodes correspond to variables, and directed edges (i.e., arrows) indicate causal influences. Additionally, the $do(x)$ operator represents the operation of setting the value of a variable $X$ to $X=x$ (i.e., the variable $X$ is instantiated to a value $x$), such that $P(y|do(x))$ represents the probability of obtaining the outcome $Y$ that would result from setting $X$ to $x$ by means of an intervention. \citet{FentonGlynn2021} proposes the following definition of actual causation\footnote{The notation and terminology follow the definition \textbf{PC1} presented by \citet[p. 72]{FentonGlynn2021}}: 
	
	\begin{probabilistic_actual_causation}
		Within a given causal model, consider a cause $X$ of an outcome $Y$ with a directed path $\mathcal{P}$ from $X$ to $Y$; let the variables that are not on $\mathcal{P}$ be denoted by $\mathbf{W}$, and the set of mediators on $\mathcal{P}$ by $\mathbf{Z}$. Given  the actually observed values $(x,\mathbf{w^*},\mathbf{z^*})$, we say that
		$X$ taking the value $X=x$ rather than $X=x'$ is the actual cause of event $Y$ (or $Y=1$) 
		when the following probability raising holds for all subsets $\mathbf{Z'}$ of $\mathbf{Z}$:
		
		\begin{equation}
			\label{eq:actual_cause_inequality}
			P(Y \,|\, do(\mathbf{W=w^*},  X=x,  \mathbf{Z'=z^*}) ) > P(Y \,|\, do(\mathbf{W=w^*},   X=x') ).
		\end{equation}
		
	\end{probabilistic_actual_causation}

	It is important to emphasize that the term actual cause refers to token causal relations, as opposed to type causal relations~\citep{Pearl2009ActualCausation,Halpern2005,FentonGlynn2017}. That is, an actual cause refers to a specific scenario, where the causal statements of the definition are regarded as \textit{singular}, \textit{single-event}, or \textit{token-level}~\citep{Pearl2009ActualCausation}. This is evident in the inequality (\ref{eq:actual_cause_inequality}), which compares the probabilities of an event or outcome $Y$ given the observed values of the variables in $\mathbf{W}$ and $\mathbf{Z'}$, which can be seen as the given context.
	
	In general, the notion of actual causation is considered the key to constructing explanations~\citep{Pearl2009ActualCausation}. \citet{Beckers2022} has used the term \textit{action-guiding explanations} for scenarios where the analysis of actual causation aims to find explanations for the outputs produced as the result of performing an action. \citet{Beckers2022} suggests that actual causes can be used for action guidance because they enable the identification of alternative actions that provide better or worse explanations of an outcome. \citet{Beckers2022} provides a conceptual analysis \footnote{The conceptual analysis conducted by \citet{Beckers2022} originally focused on notions of causal explanations in the context of deterministic causal models. Nevertheless, we consider that his conceptual accounts can be applied to probabilistic causal models.} of how action-guiding explanations relate to \textit{sufficient} and \textit{counterfactual} explanations, two other forms of potentially action-guiding explanations. In Beckers' account, a sufficient explanation indicates the conditions under which an action guarantees a particular output. On the other hand, a counterfactual explanation informs which variables would have had to be different (and in what way) for the outcome to be different. \citet[p. 2]{Beckers2022} concludes that actual causes stand between sufficient and counterfactual explanations: ``an actual cause is a part of a good sufficient explanation for which there exist counterfactual values that would not have made the explanation better."   
	
	\subsection{Using the Actual Causation Inequality to Select Alternative Actions}\label{sec:how_to_use_actual_causality}
	
	In inequality~(\ref{eq:actual_cause_inequality}), $P(Y|do(\mathbf{W=w^*}, X=x , \mathbf{Z'=z^*}) )$ provides a reference value to check for actual causation. This reference value takes into account the actual values of the variables. Using a contrastive value $X=x^\prime$ on the right side of inequality~(\ref{eq:actual_cause_inequality}), $P(Y|do(\mathbf{W=w^*}, X=x')$ is used to check whether or not probability raising holds, thereby providing a principled criterion to determine whether $X$ taking the value $X=x$ rather than $X=x'$ is the actual cause of event $Y$.
	
\begin{figure}[h!]
		\begin{center}
			\includegraphics[width=0.5\textwidth]{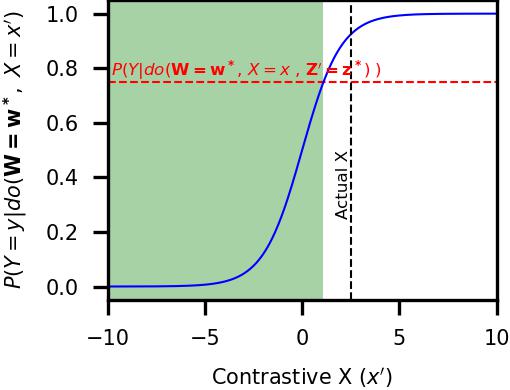}
		\end{center}
		\caption{Example of actual causation test. The curve corresponds to the right side of inequality (\ref{eq:actual_cause_inequality}) as a function of $X=x^\prime$. The horizontal line corresponds to the reference probability value (left side of inequality (\ref{eq:actual_cause_inequality}). The shaded area shows the $x^\prime$ values for which probability raising holds.   }\label{fig:example_AC_curves}
	\end{figure}

	When the variables are continuous, $P(Y|do(\mathbf{W=w^*}, X=x')$ can be plotted as a function of the contrastive value $X=x^\prime$. This is illustrated in Figure~\ref{fig:example_AC_curves}. The comparison against the reference probability $P(Y|do(\mathbf{W=w^*}, X=x , \mathbf{Z'=z^*}) )$ reveals the values $x^\prime$ for which probability raising holds (shaded region in Figure~\ref{fig:example_AC_curves}).

	Fenton-Glynn's framework of actual causation provides contrastive explanations in a given context \citep{Borner2023}. By definition, inequality (\ref{eq:actual_cause_inequality}) demands that the explanation remains valid when holding fixed all variables in $\mathbf{W}$ and (the subsets) $\mathbf{Z'}$ at their actual values; probability raising need not hold for other contexts. In this work, we propose to use the contrastive explanations obtained by applying the actual causation inequality for action guidance. Given that an outcome $Y$ has been observed, the goal is to select an alternative action that will prevent the outcome. Recalling that probability raising entails that $X$ taking the value $X=x$ rather than $X=x'$ is the cause of event $Y$, we propose to select $x'$ values as alternative actions. In the case illustrated in Figure~\ref{fig:example_AC_curves}, while selecting an $X$ value in the shaded area much smaller than the actual $X=x$ is likely to change the outcome, selecting $X$ close and above the actual value will likely leave the outcome unchanged. We emphasize that the magnitude of probability raising might drastically differ within the range of contrastive values. This has implications for the suitability of different $x\prime$ values as alternative actions. Consider the alternative actions $X=-5$ and $X=0$ in the schematic example illustrated in Figure~\ref{fig:example_AC_curves}. Based on the probabilities, it can be assumed that $X=-5$ is a better choice since selecting $X=0$ has a higher chance of leaving the outcome unchanged. Under these considerations, the selection of an alternative parameter can be based on a pre-defined probability threshold. 
	

		An alternative parameter can be found automatically with the following steps:
		\begin{enumerate}
			\item Identify the range of contrastive values where probability raising holds using inequality (\ref{eq:actual_cause_inequality}).
			\item Within these values, select the subset of values with a probability below a pre-defined probability threshold.
			\item From this second subset, select the closest value to the current parameter.
		\end{enumerate}
		In the last step we use a distance criterion. However, other application-dependent criteria can be used (e.g., the cost or availability of different alternatives) to select an alternative parameter. The crucial aspect of the automatic search is that the probability threshold is selected such that it becomes very likely to change the outcome.

	
	In summary, to check for actual causation using inequality~(\ref{eq:actual_cause_inequality}), one must decide which variable takes the role of $X$ and, given the graph structure, identify the variables for the sets $\mathbf{W}$ and $\mathbf{Z'}$. Thus, the analysis of actual causation can be applied to different variables. In the robot pouring task, if spillage occurs, the aim is to find alternative parameters to repeat the action. In this work, we use the probabilistic actual causation framework to guide the selection of alternative parameters in a principled way. In Section \ref{sec:actual_causality_results}, we illustrate in a series of detailed examples the impact of using different probability thresholds (i.e., $0.2$ for low spillage probability and and $0.5$ for chance-level probability) on the identified alternative parameters. Subsequently, in Section \ref{sec:evaluation_corrective_actions}, we evaluate the pouring success rates obtained using the alternative parameters identified using a $0.1$ probability threshold.

	\section{Related Work}\label{sec:related_work}
	This section presents robotics applications that use causal methods related to our work (i.e., actual causation, causal Bayesian networks, and causal discovery). We start by presenting applications that use the Halpern-Pearl definition of actual causality~\citep{Halpern2005}. Subsequently, we review works that use causal discovery to learn the structure of causal Bayesian networks to model robotic tasks in different contexts. Finally, we review the method proposed by \citet{Diehl2023} to predict and prevent failures using causal-based contrastive explanations, highlighting the similarities and differences to our approach.
	
	\citet{Araujo2022} use the actual causation framework by Halpern and Pearl in a human-robot interaction setting where a robot interacts with children with ASD (Autism Spectrum Disorder). The robot plays different interactive games with the children, aiming to improve the children's ability to see the world from the robot's point-of-view. The authors present a tool that uses a causal model of the interactive games and the actual causation framework. This is applied to explain events during the game's course. For example, if the robot cannot see an object involved in the interaction, it explains to the child why it cannot see it (e.g., ``I cannot see it because it is too high")~\citep{Araujo2022}. The actual cause of an event is analyzed using a rule-based system, as opposed to a search over the possible counterfactuals~\citep{Araujo2022}. The usefulness of the explanations generated by the system was evaluated by asking a group of observers to watch videos of the robot providing explanations in different situations and then rate each explanation. The rating was based on qualitative criteria, e.g., whether the explanation was understandable, sufficiently detailed, or informative about the interaction, among other aspects. 
	
	\citet{Zibaei2024} use the Halpen and Pearl actual causation framework to retrieve explanations of failure events in unmanned aerial vehicles (UAVs). In this context, an actual causation analysis aims to provide actionable explanations, that is, an explanation that indicates which corrective actions can be taken to prevent future failures. The analysis of actual failure causes was applied to different UAV failure scenarios (loss of control, events during take-off and cruise, and equipment problems). The causal models of failure events were constructed using flight logs recorded at run-time containing abstracted events and raw sensor data. In order to diagnose instances of a particular type of failure (e.g., instances of crash events in the logs), the causal graph of the failure and the actual values of the monitored event and sensor data were analyzed using a tool for the automatic checking of the Halpern and Pearl actual causation conditions. The correctness of the diagnoses was evaluated using a manually labeled ground-truth dataset.

	\citet{Chockler2021} designed an algorithm to explain the output of neural network-based image classifiers in cases where parts of the classified object are occluded. The algorithm applies concepts of the Halpern and Pearl definition of actual causation to generate explanations. The explanation consists of a subset of image pixels, which is the minimal or approximately minimal subset that allows the neural network to classify the image. It can be tested if a pixel is a cause of the classification by considering a subset of pixels in the image that does not include that pixel. In this case, applying a masking color to any combination of pixels from this subset does not alter the classification output. However, if we apply a masking color to the entire subset along with the individual pixel, this will lead to a change in the classification result. In this way, the algorithm ranks pixels according to their importance for the classification. The authors evaluate the explanations of their system by comparing their results with the outputs of other explanation tools. They compare the size of the explanation (smaller is better) and the size of the intersection of the explanation with the occluding object (smaller is better), getting favorable results for their approach. Additional work on explaining image classifiers was done by \citet{Chockler2024} and has been applied to extend the work of \citet{KommiyaMothilal2021}, who used a simplified version of the actual causation framework. The authors show that extending the previous work to the full version of the framework is possible and can benefit future work in explaining image classifiers. However, they suggest that using the full definition might make computing the explanations more complex.

	The reviewed works show that practical applications of the concepts of actual causation have focused on generating explanations, leaving the potential to guide decision-making processes aside. To the best of our knowledge, the probabilistic actual causation definition of \citet{FentonGlynn2017,FentonGlynn2021} has not been used in any practical application.

	Causal methods have been applied to model generic robot tasks such as pushing, pick-and-place, and stacking~\citep{Ahmed2020,Brawer2020,Huang2023,Diehl2023} and context-specific ones such as human-robot interaction~\cite{Castri2022} and household tasks~\citep{Li2020}. In the context of tool affordance learning, causal discovery has been used to identify the effect of push and pull actions on an object's final position in a real setup~\citep{Brawer2020}. To reduce the sim-to-real gap in robot-object trajectories, a custom causal discovery algorithm was used to optimize the simulated physical parameters~\citep{Huang2023}. Additionally, recent work has used simulations to explore large combinatoric spaces of task parameters and causal methods to model task outcomes~\citep{Ahmed2020,Huang2023,Diehl2023}.  

	Similar to our work, causal methods have been applied to find causal-based contrastive explanations for task failures~\citep{Diehl2022}. In subsequent work, this method has been extended to predict and prevent task failures by finding corrective parameters~\citep{Diehl2023}. Examples of contrastive explanations are provided for a cube stacking task and for dropping spheres into different containers (bowls, plates, and glasses)~\citep{Diehl2022}, and the method for prediction and prevention has been applied to a cube stacking task (stacking one cube and stacking three cubes)~\citep{Diehl2023}.

	In particular, the method proposed by \citet{Diehl2023} has methodological similarities to our approach. Their method uses a causal Bayesian network to predict errors and probabilities of success to find a corrective action. Simulated data are used to learn the causal Bayesian network's structure and estimate its joint probability distribution. When an action is predicted to fail, a search is conducted in a discretized parameter space to find the parametrization that needs the least interval changes to achieve a successful execution based on the predicted success probability~\citep{Diehl2023}. For example, searching for close parametrizations in the cube-stacking task means that starting from a robot hand position that will likely fail leads to finding the closest position that will likely succeed~\citep{Diehl2023}.

	The search criterion is based on contrastive explanations~\citep{Diehl2022}, which compare the variable parametrization of the failed action and the closest parametrization that exceeds a success probability threshold ($\epsilon$)~\citep{Diehl2023}. The success probabilities are retrieved from the factorized form of the joint probability distribution. The search is conducted within a tree that contains a complete parametrization of the parent variables of the outcome variable. Since the search is conducted within a tree structure, the time to find an alternative parametrization depends on the success probability threshold, the number of parent variables, and the number of discrete intervals of the variables~\citep{Diehl2023}.

	In Table~\ref{table:meta-comparison}, we summarize the similarities and differences between the method proposed by \citet{Diehl2023} and our approach. Their method automatically finds the corrective parameters, which might involve changing the value of one or more variables. In contrast, our approach is restricted to searching for corrective parameters in a single variable. Their method identifies the closest parametrization that is predicted to succeed. On the other hand, our approach identifies a range of values that explain the outcome according to the probability-raising criterion. Within this range of values, a specific value likely to yield success can be selected using a probability criterion (see explanation in Section~\ref{sec:how_to_use_actual_causality}). Similarly to~\citet{Diehl2023}, a success probability threshold can be used as a selection criterion.

	\begin{table*}
		\footnotesize
		\caption{Comparison of methods }
		\begin{center}
			\begin{tabular}{C{1.2cm}C{2.0cm}C{1cm}C{1cm}C{1.5cm}C{2.6cm}C{0.05cm}C{1cm}C{1.8cm}C{1.5cm}}
				\hline
				& \multicolumn{5}{c}{\textbf{Similarities}} & & \multicolumn{3}{c}{\textbf{Differences}}  \\
				\cline{2-6} \cline{8-10}
				& \textbf{Model} & \textbf{Training data} &\textbf{Type of explanation} & \textbf{Selection of corrective action} & \textbf{Input} & & \textbf{Parameter space} & \textbf{Corrected variables} & \textbf{Contrasted probabilities} \\
				\cline{1-6} \cline{8-10}
				\cite{Diehl2023} & Causal Bayesian network (discrete variables)& Simulation & Contrastive & Based on a probability threshold & Current values of all variables (i.e., parametrization) and probability of the outcome variable & & Discrete & The corrective parametrization might involve changing one or more variables & Conditional probabilities from the factorized joint probability distribution \\
				\cline{1-6} \cline{8-10}
				Ours & Causal Bayesian network (continuous and discrete variables)& Simulation & Contrastive & Based on a probability threshold & Current values of all variables (i.e., parametrization), causal probability (do-operator) of the outcome variable, and variable to correct & & Continuous & The corrective parametrization changes one variable & Causal probabilities (do-operator)  \\
				\hline
			\end{tabular}
		\end{center}
		\label{table:meta-comparison}
	\end{table*}
	
	\section{Materials and Methods}\label{sec:materials_and_methods}
	\subsection{Robot Pouring Simulation}\label{sec:robot_simulation}
\begin{figure}[h]

		\centering
		\begin{subfigure}[b]{0.22\textwidth}
			\centering
			\includegraphics[width=\linewidth]{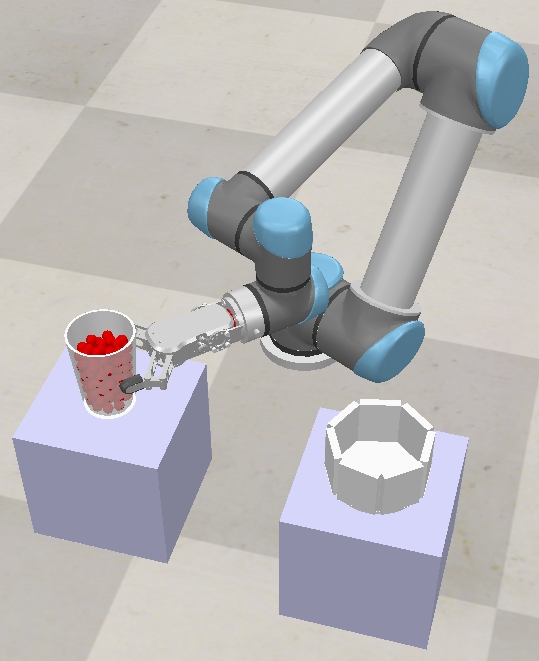}
			\caption{Trial start}
			\label{fig:trial_start}
		\end{subfigure}  
		\begin{subfigure}[b]{0.22\textwidth}
			\includegraphics[width=\linewidth]{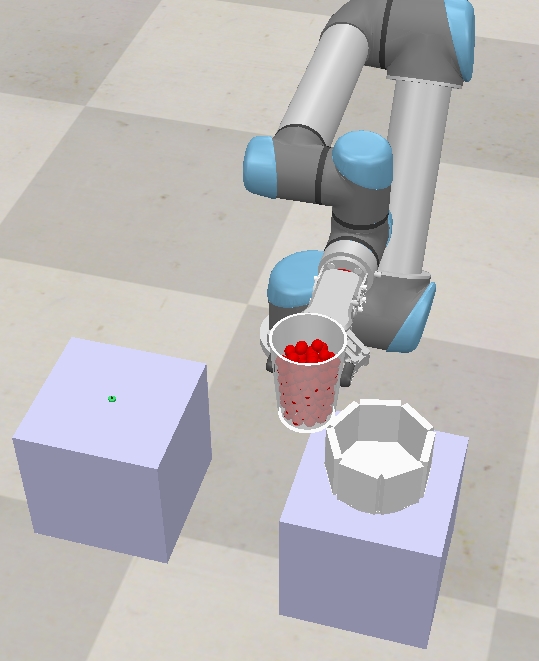}
			\caption{Transport}
			\label{fig:trial_pouring_position}
		\end{subfigure}
		\begin{subfigure}[b]{0.22\textwidth}
			\includegraphics[width=\linewidth]{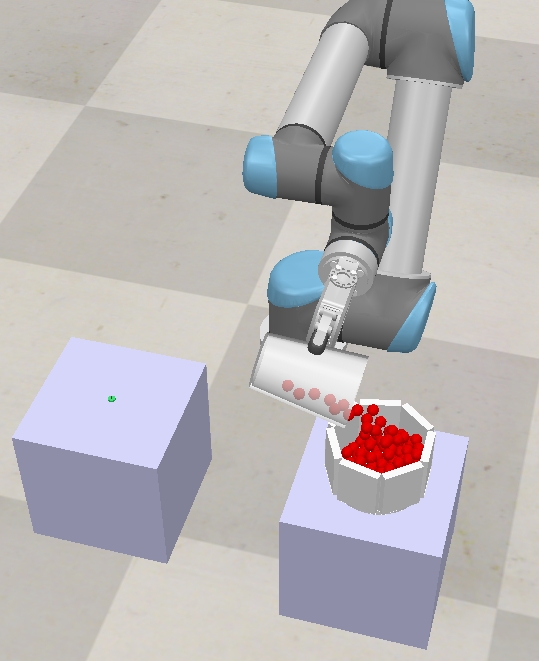}
			\caption{Pouring}
			\label{fig:trial_pouring}
		\end{subfigure}
		\begin{subfigure}[b]{0.22\textwidth}
			\includegraphics[width=\linewidth]{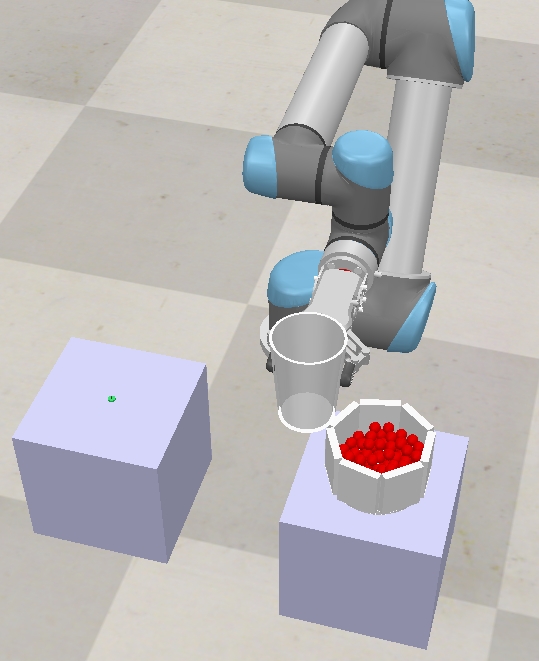}
			\caption{Trial end}
			\label{fig:trial_end}
		\end{subfigure}
		\caption{Course of pouring trial. \textbf{(a)}~Source container is filled with simulated marbles and a target container of random dimensions is generated. \textbf{(b)}~The source container is transported to the pouring position. \textbf{(c)}~The source container is rotated to pour the marbles into the target container. \textbf{(d)}~Trial end.}	\label{fig:pouring_simulation}
	\end{figure}

	The robot pouring task was simulated using CoppeliaSim Version 4.8.0 (rev. 0)~\cite{CoppeliaSim} with the Open Dynamics Engine (ODE), a physics engine for rigid body dynamics and collision detection. The setup consists of an UR5 robot arm with a parallel jaw gripper, as shown in Figure~\ref{fig:pouring_simulation}. In the simulation, the robot poured marbles from a source container into a target container. The marbles were simulated with CoppeliaSim's particle object, which simulates spherical particles using parameters for their diameter ($1.5$~$cm$) and density ($2829.42$ $kg/m^3$, empirically determined).

	In each trial, the source container (capacity = $514.72$ $cm^3$ ) was filled with simulated marbles (Figure~\ref{fig:trial_start}). The characteristics of the pouring movement were fixed for all trials. The robot grasped the source container and brought it to a pouring position relative to the target container's rim (Figure~\ref{fig:trial_pouring_position}). In the pouring position, the gripper was rotated with a fixed rotation velocity ($1$ $rad/s$) until reaching a fixed angle of $-15^{\circ}$ (Figure~\ref{fig:trial_pouring}). After pouring all the marbles (Figure~\ref{fig:trial_end}), spillage was detected using a force sensor located on the base of the target container. In each trial we manipulated the amount of marbles in the source container and the dimensions of the target container. The dimensions of the source container remained fixed across trials. In each trial, the source container was filled with a random amount of marbles, and a target container with random dimensions (height and rim diameter) was generated. The positioning of the marbles inside the source container was random. Six thousand pouring trials were simulated. The rationale of the random trial parameters and variable definitions are explained in Section \ref{sec:variables}. Details of the random distributions used to sample the trial parameters are provided in Table~\ref{table:variables}. In each trial, the randomness of the spillage outcome results from the behavior of the particles simulated by the physics engine and its interaction with the characteristics of the target container.
	
	\subsection{DAG Variables}\label{sec:variables}
	In each trial, the source container was filled to a random fullness level, and a target container with random dimensions was generated. The rationale for the poured amounts and the target container characteristics in each trial aims to capture typical spillage causes. Table \ref{table:variables} provides details of the variable definitions and random sampling distributions. The target container can have an equal, smaller, or larger capacity than the source container. The capacity of the target container influences the probability of spillage (spillage is more likely to occur when pouring into a target container of a small capacity). We represent this cause of spillage with the variable \textit{relative capacity} ($RC$). 
	
	Aside from the target container's capacity, spillage depends on the complex interplay between the fullness level of the source container and the rim dimensions of the source and target containers. Imagine pouring marbles from a wide glass into a bottle through its narrow mouth. While pouring a single marble is likely to succeed, the amount spilled will increase as the number of marbles increases. We represent the fullness level of the source container with the variable \textit{fullness} ($FU$).
	
	The target container's rim can be equal, smaller, or larger in diameter than the source container. The relation between rim diameters influences the probability of spillage (spillage is more likely to occur when pouring into a target container with a smaller rim diameter). We represent this cause of spillage with the variable \textit{relative diameter} ($RD$). As a result, in each trial, the dimensions of the target container (height and rim diameter) are determined by the $RC$ and $RD$ variables. 
	
	We also consider the fact that spillage occurs when the amount of marbles exceeds the capacity of the target container. The source containers' capacity and fullness level determine the poured amount (recall that all the marbles are poured). We represent the difference in volume between the poured amount and the target container's capacity with the variable \textit{relative volume} ($RV$). Finally, we represent the outcome of the pouring trial with the variable \textit{spillage} ($S$), a binary variable that indicates whether or not spillage occurred. $S$ is labeled as $true$ irrespective of the number of spilled marbles (i.e., spilling one or twenty marbles yields $S=true$).
	
	\begin{table*}
		\caption{Variable definitions and sampling of trial parameters }
		\begin{center}
			\begin{tabular}{ m{1.3cm} m{1cm} m{12cm} }
				\hline
				\multicolumn{1}{l}{\rule{0pt}{12pt}
					Variable}&\multicolumn{1}{c}{\rule{0pt}{12pt}
					Acronym}&\multicolumn{1}{l}{Definition and sampling distribution}\\[2pt]
				\hline\rule{0pt}{12pt}
				
				\textit{relative capacity} &$RC$ & $RC$ represents the relation between the containers' capacities, defined as  $RC=\frac{target \ capacity}{source \ capacity}$. Sampled from a truncated Gaussian distribution $ \mathcal{N}_{trunc.} (\mu=1.0, \sigma=0.25, min=0.5, max=2.0)$. $RC<1$ corresponds to target containers of lower capacity, and $RC>1$ to target containers of larger capacity.\\[1.2cm]
				
				\textit{fullness} & $FU$ & $FU$ expresses the fullness level of the source container as a fraction. Sampled from $\mathcal{N}_{trunc.} (\mu=0.7, \sigma=0.2, min=0.3, max=1.0)$. For example,  $FU=0.5$ corresponds to a half-full source container. \\
				
				\textit{relative diameter} & $RD$ & $RD$ represents the relation between the container rim diameters, defined as  $RD=\frac{target \ diameter}{source \ diameter}$. Sampled from $ \mathcal{N}_{trunc.} (\mu=1.0, \sigma=0.25, min=0.5, max=1.5)$. $RD<1$ corresponds to target containers of smaller diameter, and $RD>1$ corresponds to target containers of larger diameter. The target container's height and diameter were determined based on the sampled $RC$ and $RD$.\\
				
				\textit{relative volume} & $RV$ & $RV$ represents the volume relation between the poured amount and the target container's capacity, defined as $RV=\frac{poured \ volume}{target \ capacity}$.  $RV<1$ indicates that the poured amount fits into the target container, and $RV>1$ indicates that the poured amount exceeds the target container's capacity.\\
				
				\textit{spillage} & \textit{S} & Binary variable indicating whether or not spillage occurred. \\[2pt]
				\hline
			\end{tabular}
		\end{center}
		\label{table:variables}
	\end{table*}
	
	\subsection{Determination of DAG Structure Using Causal Discovery}\label{sec:discovery}
	In order to perform an analysis of actual causation, we require a DAG of the causes of spillage, where edges represent probabilistic effects between variables. A naive approach to setting the causal structure would be to assume that each variable described in Section~\ref{sec:variables} is connected to the outcome $S$ with an edge. This would constitute a strong assumption in which all variables are direct causes of $S$, excluding the possibility of indirect effects. In this respect, it is important to recall that the validity of the analysis of actual causation relies on the correctness of the DAG structure. To avoid making naive assumptions about the causal structure, we leverage the availability of simulated pouring trials to determine the structure of the DAG using a causal discovery algorithm. 
	
\begin{figure}[h!]
		\begin{center}
			\includegraphics[width=0.25\textwidth]{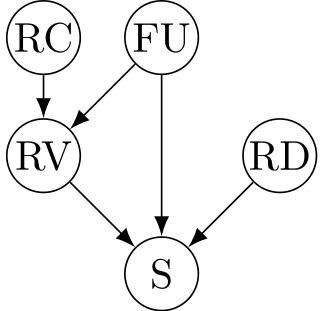}
		\end{center}
		\caption{Discovered DAG}\label{fig:dag}
	\end{figure}

	In order to determine the structure of the DAG, the variables described in Section \ref{sec:variables} were processed with the PC algorithm~\citep{Spirtes1991}, a well-established causal discovery algorithm~\citep{Glymour2019,Nogueira2022}. In general, causal discovery algorithms, also known as structure learning algorithms, perform a systematic analysis of many possible causal structures, typically by testing probabilistic independence and dependence between the variables~\citep{MalinskyDanks2017,Glymour2019,Nogueira2022}. We used the PC implementation available in Tetrad (version 7.6.5-0)\footnote{Publicly available at: \url{https://www.ccd.pitt.edu/tools/} , access: 06.12.24}, a software toolbox for causal discovery~\citep{tetrad}. As a statistical test, we use the Degenerate Gaussian Likelihood Ratio Test (DG-LRT)~\citep{AndrewsEtal2019DG}, which has shown good discovery performance on datasets containing continuous and discrete variables~\citep{AndrewsEtal2019DG}. We executed the algorithm on 1000 bootstraps of the data to ensure the stability and reliability of the inferred causal relationships (if the results vary widely over the different bootstrap samples, the output of the algorithm is considered unstable)~\citep{Glymour2019}. Further details about the PC parameters, bootstrapping results, and modeling assumptions are provided in the supplementary materials. The discovered DAG is shown in Figure~\ref{fig:dag}.
	
	\subsection{Estimation of do-Probabilities Using Neural Autoregressive Density Estimation}\label{sec:NADE}
	In this paper, we use \textit{neural autoregressive density estimators} (NADEs)~\citep{Garrido2021} to estimate the conditional distributions necessary to compute the interventional queries in inequality (\ref{eq:actual_cause_inequality}). NADEs are used as universal approximators and are flexible to work with models containing continuous and discrete variables~\citep{Garrido2021}. \citet{Garrido2021} demonstrated in a series of examples that NADEs provide a practical modeling architecture to estimate causal quantities from models with linear and non-linear relationships between variables of different distributions (e.g., Normal, Log-normal, and Bernoulli) using the DAG formalism and the do-calculus framework. 
	
	A DAG model represents the conditional dependencies between the set of $J$ random variables $X_1 , \cdots, X_J$. The DAG induces a joint probability distribution $P(X)$ of the variables, which can be factorized into the 
	conditional distributions of each $X_j$ conditioned on a function $f_j$ of its parents $PA(X_j)$ (the causal Markov condition):
	
	\begin{equation}\label{eq:factorized_joint_probability}
		P(X) = \prod_{j}^{J} P(X_j | f_j( PA (X_j) ))
	\end{equation}
	
	A NADE is a generative model that estimates the conditional distributions in equation (\ref{eq:factorized_joint_probability})~\citep{Garrido2021}. The functions $f_j$ are parametrized as independent fully connected feed-forward neural networks. Therefore, there is a neural network for each variable in the DAG. Each neural network takes the parents of the variable of interest $PA(X_j)$ and outputs the parameters of the distribution of $X_j$. For example, the neural network outputs the mean and standard deviation of Gaussian variables. These networks are trained using the negative log-likelihood of equation (\ref{eq:factorized_joint_probability}) as the loss function. The individual conditional distributions in equation (\ref{eq:factorized_joint_probability}), also called independent causal mechanisms or Markov Kernels, are used to estimate the effects of interventions~\citep{Garrido2021}. The causal estimates are reliable under the assumption that the DAG structure is correct and that the training data provides enough support to learn the distribution parameters~\citep{Garrido2021}. Details of the implementation are provided in the supplementary material.
	
	\subsection{Causal Probability Expressions and Actual Causation Inequalities}\label{sec:dag_and_formulas}
	The factorized joint probability distribution that results from the DAG structure shown in Figure~\ref{fig:dag} is expressed as a product of conditional distributions and independent causal mechanisms:
	
	\begin{equation}\label{eq:joint_prob_dist}
		P(RC,FU,RV,RD,S) = P(RC)\cdot P(FU)\cdot P(RV|RC,FU)\cdot P(RD)\cdot P(S| FU, RD, RV)		
	\end{equation}
	
\begin{figure}[h!]
		\begin{center}
			\includegraphics[width=\textwidth]{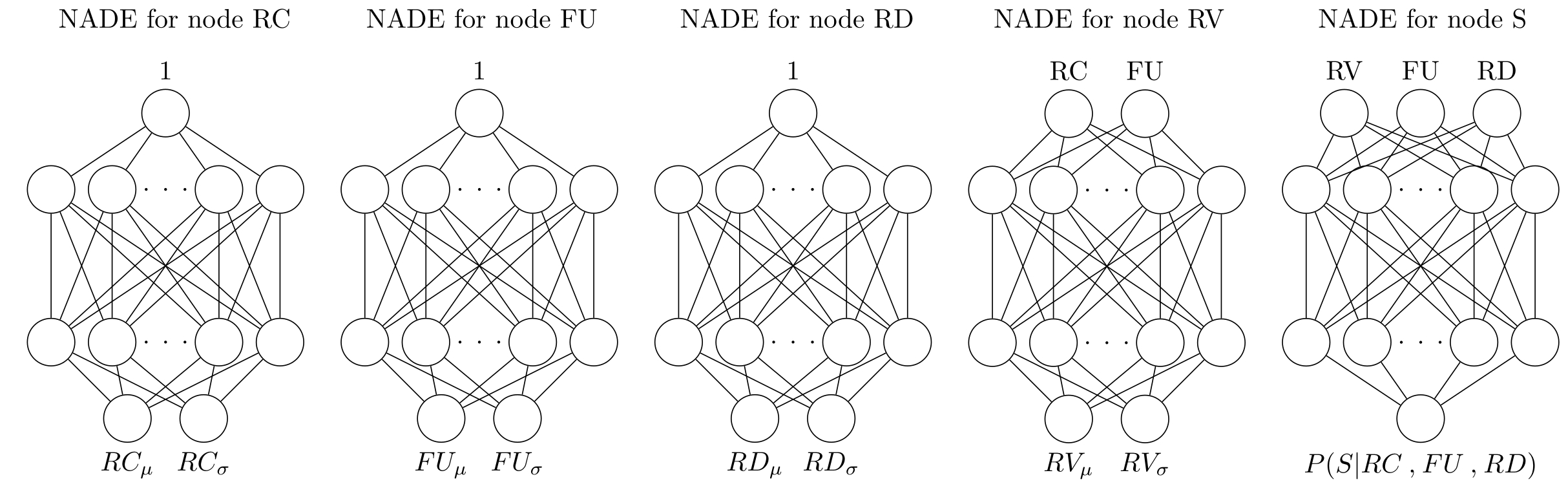}
		\end{center}
		\caption{Implemented NADEs for the DAG nodes and auxiliary NADE used for causal effect estimation}\label{fig:nades}
	\end{figure}

	These conditional distributions and independent causal mechanisms are approximated using NADEs, as described in Section~\ref{sec:NADE}. The implemented neural networks are shown in Figure~\ref{fig:nades}. The distribution of the continuous variables $RC$, $FU$, $RV$, and $RD$ is approximated by Gaussian distributions. For these variables, each network takes the parent of the corresponding variable as input and produces two parameters, one for the mean ($\mu$) and one for the standard deviation ($\sigma$) of the Gaussian distribution. Following the implementation of \citet{Garrido2021}, the root nodes ($RC$, $FU$, and $RD$) take a constant value as input, represented as a ``1" in Figure~\ref{fig:nades}. The discrete variable $S$ follows a Bernoulli distribution. Its neural network takes the variables $RV$, $FU$, and $RD$ as input and outputs the probability of sampling $S=true$ given the input values. The NADE estimators are used to compute causal (or interventional) probabilities (e.g., $P(S|do(RD))$ ) and actual causation queries in the form of inequality~(\ref{eq:actual_cause_inequality}).

	\subsubsection{Causal Probability Expressions}
	The DAG indicates that $S$ has direct and indirect causes. The path $RC\rightarrow RV \rightarrow S$ shows that $RC$ is an indirect cause of $S$, and its effect is mediated by $RV$. The causal relation between $FU$ and $S$ combines a direct path ($FU \rightarrow S$) and indirect path ($FU \rightarrow RV \rightarrow S$). Finally, $RD$ is a direct cause of S (path $RD\rightarrow S$). Ultimately, the probability of $S$ depends on the interactions between $RV$, $FU$, and $RD$. Additional insight into the factors that yield spillage can be gained by estimating the effects of each variable on $S$. Causal effects are functions of  interventional probabilities $P(S|do(X))$ over different values of $X$; the expressions can be obtained using the rules of $do$-calculus with the \textit{causaleffect} R package (version 1.3.15) \citep{Rcausaleffect}. Following the procedure done by \citet{Garrido2021}, we use Monte Carlo integration in order to obtain a numerical approximation of the causal expressions of interest.

	Under the given causal model, the interventional probability  of spillage $P(S|do(RC))$  under a given choice of $RC$ is obtained as:

	\begin{equation}\label{eq:do_cd}
		\begin{split}
			P(S\mid do(RC)) & = \int_{RD,FU,RV} P(S | FU, RD, RV) \cdot P(RV | RC, FU) \cdot P(RD) \cdot P(FU)\, dRV \, dRD \, dFU\\
		\end{split}
	\end{equation}
	
	We approximated $P(S|do(RC))$ using Monte Carlo integration by sampling from the distributions $P(RD)$, $P(FU)$, and $P(RV | RC, FU)$, and propagating forward through the neural network implemented for $P(S | FU, RD, RV)$.   
	
	The interventional probability $P(S|do(FU))$ for a given choice of $FU$ is obtained as:
	
	\begin{equation}\label{eq:do_fu}
		\begin{split}
			P(S\mid do(FU)) &= \int_{RC,RV,RD} P(S|FU,RD,RV) \cdot P(RV|RC,FU) \cdot P(RD) \cdot P(RC) \, dRC \, dRD \, d RV\\ 
		\end{split}
	\end{equation}
	
	Similarly, $P(S|do(FU))$  was approximated using Monte Carlo integration by sampling from the distributions $P(RC)$, $P(RD)$, and $P(RV | RC, FU)$, and propagating forward through the neural network implemented for $P(S | FU, RD, RV)$.  
	
	The expression $P(S|do(FU))$ in equation (\ref{eq:do_fu}) combines the effect mediated by $RV$ and the direct effect of $FU$. In order to assess the direct effect of $FU$ on $S$ we must fix $RV$. This enables us to assess how the direct effect of $FU$ may differ for different choices of $RV$. The direct effect can be expressed by varying the values of $FU$ with constant $RV$ in $P(S|do(FU,RV))$, which is obtained as: 
	
	\begin{equation}\label{eq:do_direct_fu}
		\begin{split}
			P(S\mid do(FU,RV)) &= \int_{RD}  P\left(S| RD,FU,RV\right)\cdot P\left(RD\right) \, dRD \\
		\end{split}
	\end{equation}
	
	The probability $P(S|do(FU,RV))$ 
	was also approximated using Monte Carlo integration by sampling from the distributions $P(RD)$ and propagating forward through the neural network implemented for $P(S | FU, RD, RV)$.

	Furthermore, $P(S|do(RV))$ is obtained as:
	
	\begin{equation}\label{eq:do_vd}
		\begin{split}
			P(S\mid do(RV)) &= \int_{RD,FU} P(S|RD,FU,RV) \cdot P\left(RD\right)  \cdot P(FU)\, dFU\,dRD 
		\end{split}
	\end{equation}
	
	The probability $P(S|do(RV))$ was approximated using Monte Carlo integration by sampling from the distributions $P(RD)$ and $P(FU)$, and propagating forward through the neural network implemented for $P(S | RD, FU, RV)$.  
	
	Finally, $P(S|do(RD))$ is obtained as:
	\begin{equation}\label{eq:do_dd}
		\begin{split}
			P(S\mid do(RD)) &= \int_{RC,FU,RV} P(S|FU,RD,RV) \cdot P(RV|RC,FU) \cdot P(FU) \cdot P(RC) \, dRC \, dFU \, dRV \\
		\end{split}
	\end{equation}
	
	The probability $P(S|do(RD))$ was approximated using Monte Carlo integration by sampling from the distributions $P(FU)$, $P(RC)$, and $P(RV | RC, FU)$, and propagating forward through the neural network implemented for $P(S | FU, RD, RV)$.   
	
	\subsubsection{Inequalities for Analysis of Probabilistic Actual Causation}
	For the analysis of actual causation we identify the sets of variables necessary to compute the probabilities compared in inequality (\ref{eq:actual_cause_inequality}). In order to analyze whether $RD$ is an actual cause of $S$, we consider the only path $\mathcal{P}$: 
	$RD \rightarrow S$. The set of variables $\mathbf{W}$ that lie off the path is $\mathbf{W} = \{RC,FU,RV\}$, and the set $\mathbf{Z}$ of variables that lie intermediate between $RD$ and $S$ is $\mathbf{Z} = \emptyset$. $RD$ taking the value $RD=rd$ rather than $RD=rd'$ (in the context of the observed values $rc,fu,rv$) is an actual cause of $S$ 
	(or $S=true$) when the following probability raising holds:
	
	\begin{equation}\label{eq:dd_actual_cause_inequality}
		\begin{split}
			P(Y\mid do(\mathbf{W=w^*}, X=x , \mathbf{Z'=z^*}) ) & > P(Y\mid do(\mathbf{W=w^*}, X=x') ) \\		
			P( S\mid do(rc,fu,rv,rd)) & > P( S \mid do(rc,fu,rv,rd')) \\
			P\left(S\mid fu,rv,rd\right) & > P\left(S\mid fu,rv,rd'\right) 
		\end{split}
	\end{equation}
	
	The conditional probabilities in inequality (\ref{eq:dd_actual_cause_inequality}) were approximated using the neural network for $P(S|RV,FU,RD)$.
	
	In order to analyze whether $FU$ is an actual cause of $S$, we consider  the path $\mathcal{P}$: 
	$FU \rightarrow RV \rightarrow S$. The set of variables $\mathbf{W}$ that lie off the path is $\mathbf{W} =\{RC,RD\}$, and the set $\mathbf{Z}$ of variables that lie intermediate between $FU$ and $S$ on $\mathcal{P}$ is $\mathbf{Z} =\{ RV \}$. Recalling that inequality (\ref{eq:actual_cause_inequality}) is tested for all the subsets of $\mathbf{Z}$ which includes the empty set~\citep{FentonGlynn2021},  $FU$ taking the value $FU=fu$ rather than $FU=fu'$ is an actual cause of $S$ (or $S=true$) when the probability raising holds both for $\mathbf{Z'}=RV$ (inequality (\ref{eq:fu_actual_cause_inequality})) and $\mathbf{Z'}=\emptyset$ (inequality (\ref{eq:fu_empty_z_actual_cause_inequality})) :
	
	\begin{equation}\label{eq:fu_actual_cause_inequality}
		\begin{split}
			P(Y\mid do(\mathbf{W=w^*}, X=x, \mathbf{Z'=z^*}) ) & > P(Y\mid do(\mathbf{W=w^*}, X=x') ) \\		
			P( S\mid do(rc,rd,fu,rv)) & > P( S\mid do(rc,rd,fu')) \\
			P\left(S\mid rd,fu,rv\right) & >  \int_{RV} P\left(S\mid rd,fu',RV\right) \cdot P\left(RV\mid rc,fu'\right) dRV			
		\end{split}
	\end{equation}
	
	\begin{equation}\label{eq:fu_empty_z_actual_cause_inequality}
		\begin{split}
			P(Y\mid do(\mathbf{W=w^*}, X=x, \mathbf{Z'=z^*}) ) & > P(Y\mid do(\mathbf{W=w^*}, X=x') ) \\		
			P( S\mid do(rc,rd,fu)) & > P( S\mid do(rc,rd,fu')) \\
			\int_{RV} P\left(S\mid rd,fu,RV\right) \cdot P\left(RV\mid rc,fu\right) dRV	& >  \int_{RV} P\left(S\mid rd,fu',RV\right) \cdot P\left(RV\mid rc,fu'\right) dRV	
		\end{split}
	\end{equation}
	
	The integrals in inequality (\ref{eq:fu_empty_z_actual_cause_inequality}) and on the right side of inequality (\ref{eq:fu_actual_cause_inequality}) were approximated using Monte Carlo integration by sampling from the distribution $P(RV\,|\, RC,FU)$ and propagating forward through the neural network implemented for $P(S \,|\, FU, RD, RV)$.  
	
	\section{Results}
	\subsection{Causal Probabilities}\label{sec:causal_effect_results}
\begin{figure}

		\centering
		\begin{subfigure}[b]{0.32\textwidth}
			\includegraphics[width=\linewidth]{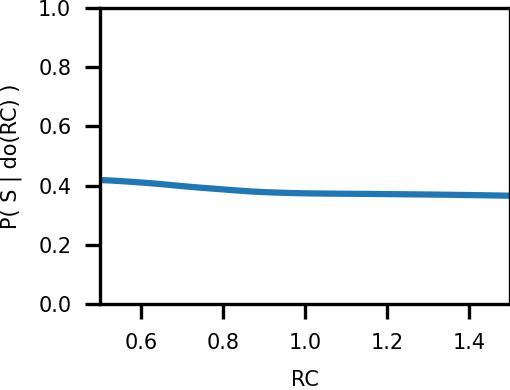}
			\caption{Causal probability of $RC$ on $S$.}
			\label{fig:doCD}
		\end{subfigure}  
		\begin{subfigure}[b]{0.32\textwidth}
			\includegraphics[width=\linewidth]{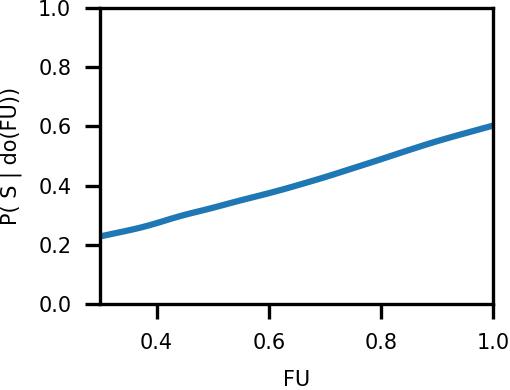}
			\caption{Total causal probability of $FU$ on $S$.}
			\label{fig:doFU}
		\end{subfigure}
		\begin{subfigure}[b]{0.32\textwidth}
			\includegraphics[width=\linewidth]{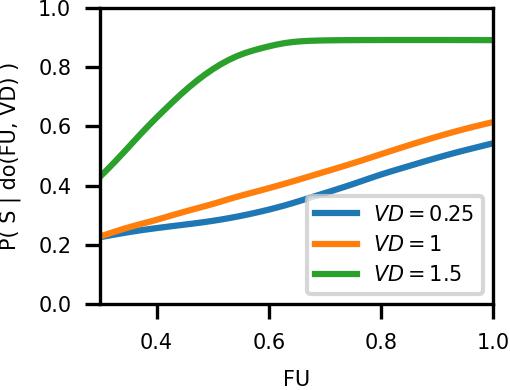}
			\caption{Direct causal probabilty of $FU$ on $S$.}
			\label{fig:doFUdirect}
		\end{subfigure}
		\\\vspace{0.5cm}
		\begin{subfigure}[b]{0.3\textwidth}
			\includegraphics[width=\linewidth]{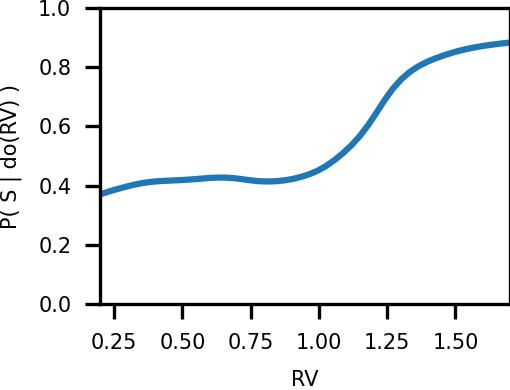}
			\caption{Causal probability of $RV$ on $S$.}
			\label{fig:doVD}
		\end{subfigure}
		\begin{subfigure}[b]{0.3\textwidth}
			\includegraphics[width=\linewidth]{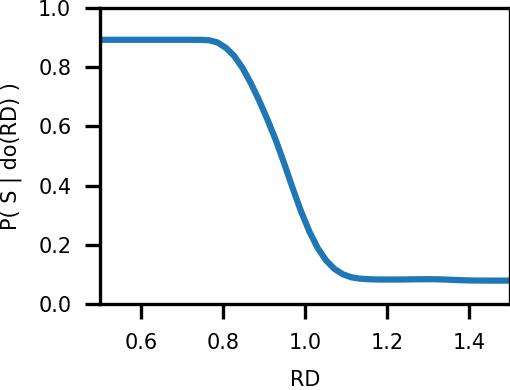}
			\caption{Causal probability of $RD$ on $S$.}
			\label{fig:doDD}
		\end{subfigure}

		\caption{Causal probabilities of DAG variables.}	\label{fig: subfigures}
	\end{figure}

	The causal probability $P(S|do(RC))$ was computed using equation (\ref{eq:do_cd}) over a range of RC values covering low ($RC<1$) and large ($RC>1$) target container capacities. The estimated probabilities are shown in Figure~\ref{fig:doCD}. Over the whole range of $RC$, the probability lies slightly below 0.5. This probability curve reflects the fact that $RV$ mediates the effect of $RC$, which in turn crucially depends on $FU$. Without knowing the context of $FU$, the total effect of $RC$ on $S$ 
	lies around the chance level. Thus, solely knowing or setting a $RC$ value does not provide enough information on causing or preventing spillage.

	The probabilities $P(S|do(FU))$ and $P(S|do(FU, RV))$ were computed using equations~(\ref{eq:do_fu}) and~(\ref{eq:do_direct_fu}), respectively, over a range of $FU$ values covering low ($FU<0.4$), medium ($0.4 \leq FU \leq 0.6$), and high ($FU>0.6$) fullness levels of the source container. The estimated probabilities for $P(S|do(FU))$, shown in Figure~\ref{fig:doFU}, indicate that low fullness levels yield a low probability ($\approx 0.2$) of spillage. This probability increases as the fullness level increases, reaching a maximum of $0.6$. For the direct effect $P(S|do(FU, RV))$, we present the probabilities obtained for three levels of $RV$: $RV=0.25$ (the poured amount fits into the target container), $RV=1$ (the poured amount equals the target container's capacity), and $RV=1.5$ (the poured amount exceeds the target container's capacity). The estimated probabilities are shown in Figure~\ref{fig:doFUdirect}. It can be observed that the probabilities obtained for $RV=0.25$ and $RV=1$ are similar. Thus, for $RV<1$, increasing $FU$ will have a similar effect on the probability of spillage. In contrast, the results obtained for $RV=1.5$ show different non-linear behavior with overall spillage probabilities above the chance level.         
	
	The probability $P(S|do(RV))$ was computed using equation~(\ref{eq:do_vd}) over a range of $RV$ covering values where the poured amount fits into the target container ($RV<1$) and the poured amount exceeds the target container's capacity ($RV>1$). The estimated probabilities are shown in Figure~\ref{fig:doVD}. For $RV<1$, we obtained spillage probabilities $\approx 0.4$. Therefore, without knowing the context of the other variables, when $RV<1$, the probability of spillage lies slightly below the chance level. For $RV>1$ we observe a moderate probability increase, reaching a maximum of $0.88$. By definition, $RV>1$ indicates that the poured amount exceeds the target container's capacity. Thus, in practice, a large probability of spillage~($\approx 1$) for $RV \gg 1$, and a step probability increase for $RV \approx 1$ would have been expected. This suggests that the NADE for $RV$ is smoothing the estimated probabilities.

	The probability $P(S|do(RD))$ was computed using equation~(\ref{eq:do_dd}) over a range of $RD$ values covering target containers of smaller diameter ($RD<1$) and target containers of larger diameter ($RD>1$). The estimated probabilities are shown in Figure~\ref{fig:doDD}. For $RD<0.8$, we observe a constant large probability of spillage~($\approx 0.9$). For $0.8\leq RD \leq 1.1$, we observe a sharp probability decrease. For $RD>1.1$ we observe a low probability of spillage.   
	
	In contrast to the results obtained for $P(S|do(FU))$ and $P(S|do(RV))$,  $P(S|do(RD))$ shows distinct ranges where $RD$ yields low and high probability of spillage. Additionally, the range of values where $P(S|do(RD))$ shows a sharp probability decrease indicates that a slight change in $RD$ can significantly affect the probability of spillage.

	Overall, the causal effects presented in this section exhibit non-linear behavior. Due to the non-linear behavior and the different magnitudes of the spillage probability, using the individual causal probabilities to analyze a trial with particular $RC$, $FU$, $RV$, and $RD$ values does not provide conclusive information. This happens because causal probabilities in the form of $P(S| do(X) )$ do not consider the context (i.e., the actual value) of the other variables.  In the following section, we present a set of examples where the actual causation framework is used to determine which variable (at its actual value) is an actual cause of spillage and how the variable should be changed to yield a different outcome.
	
	\subsection{Actual Causes of Spillage and Alternative Actions}\label{sec:actual_causality_results}
	As explained in Section~\ref{sec:actual_causality_framework}, the actual causation inequality compares a reference probability value against the probabilities obtained from contrastive values to check whether or not probability raising holds. As proposed in Section~\ref{sec:how_to_use_actual_causality}, we use the contrastive values where probability raising holds to guide the selection of alternative parameters to avoid spillage. In this section, we present the results obtained by using the automatically-selected alternative values to achieve either a chance-level or a low probability of spillage. In the following, we explain the rationale behind the selected spillage trial examples. 
	
	According to the considerations presented in Section~\ref{sec:variables} and the DAG discovered from the training data (Figure~\ref{fig:dag}), spillage has three direct causes: $RV$, $FU$, and $RD$. At the beginning of the pouring action, i.e., at the pouring onset, the interaction of $FU$ and $RD$ determines the probability of spillage. Afterward, as the content is poured into the target container, spillage will occur after the poured amount exceeds the container's capacity; that is, spillage caused by $RV$. Recalling that $RV$ is caused by $FU$ and $RC$, the capacity of the target will be exceeded only if $RC<1$. The smaller the $RC$, the smaller the $FU$ will be necessary to exceed the target container's capacity. Otherwise, when $RC>1$, spillage is caused only by $FU$ and $RD$. Therefore, we exemplify the usage of the actual causation framework for the analysis of spillage trials with $0.5<RC<1$. For simplicity, we focus the analysis on $RD$ and $FU$, which have a direct effect on $S$. We do not consider $RC$ because it doesn't have a direct effect on $S$ (the effect of $RC$ is mediated by $RV$, which also depends on $FU$).

	In each example, we start by presenting the frequency of the outcomes (spillage true or false) over 100 replications using the actual trial parameters. Afterward, we perform the actual causation analysis of the $FU$ and $RD$ variables. To compare the suitability of different alternative parameters, we select alternative $FU$ or $RD$ parameters where $P(Y|do(\mathbf{W=w^*}, X=x')$ yields a low probability (0.2) and a chance-level (0.5) probability. The alternative parameters are identified automatically following the steps described in Section \ref{sec:how_to_use_actual_causality}. We conduct 100 replications using the alternative parameters and present the frequency of the outcomes.  
	
	\subsubsection{Example 1}
\begin{figure}[h]
		\centering
		\begin{subfigure}[t]{0.24\textwidth}
			\centering
			\includegraphics[width=\linewidth]{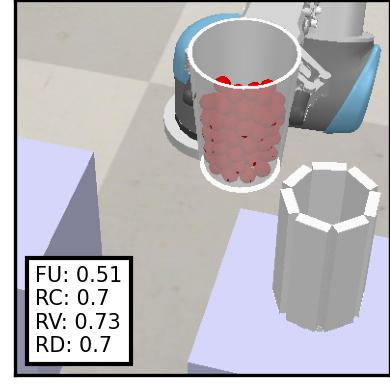}
			\caption{Actual trial parameters}
			\label{fig:actual_e1}
		\end{subfigure}  
		\begin{subfigure}[t]{0.24\textwidth}
			\includegraphics[width=\linewidth]{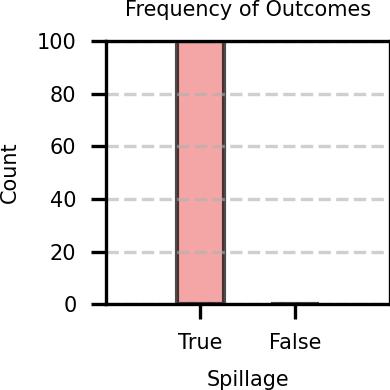}
			\caption{Outcomes with actual parameters}
			\label{fig:repli_actual_e1}
		\end{subfigure}
		\begin{subfigure}[t]{0.24\textwidth}
			\includegraphics[width=\linewidth]{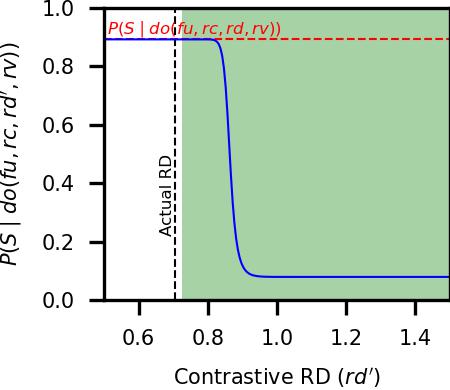}
			\caption{$RD$ as actual cause}
			\label{fig:ac_dd_e1}
		\end{subfigure}
		\begin{subfigure}[t]{0.24\textwidth}
			\includegraphics[width=\linewidth]{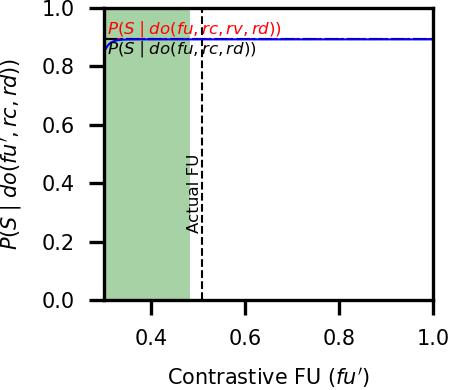}
			\caption{$FU$ as actual cause}
			\label{fig:ac_fu_e1}
		\end{subfigure}
		\\\vspace{0.5cm}
		\begin{subfigure}[t]{0.24\textwidth}
			\centering
			\includegraphics[width=\linewidth]{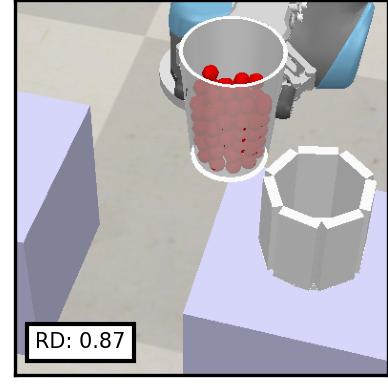}
			\caption{Alternative $RD$ }
			\label{fig:alt_dd_0p5_e1}
		\end{subfigure}
		\begin{subfigure}[t]{0.24\textwidth}
			\centering
			\includegraphics[width=\linewidth]{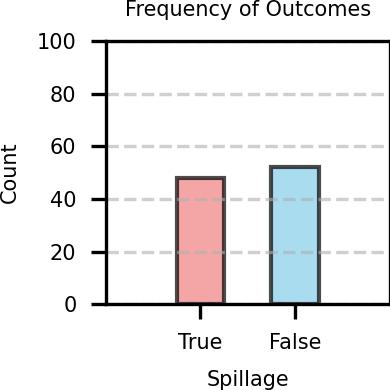}
			\caption{Outcomes with alternative $RD$}
			\label{fig:repli_alt_dd_0p5_e1}
		\end{subfigure}
		\begin{subfigure}[t]{0.24\textwidth}
			\centering
			\includegraphics[width=\linewidth]{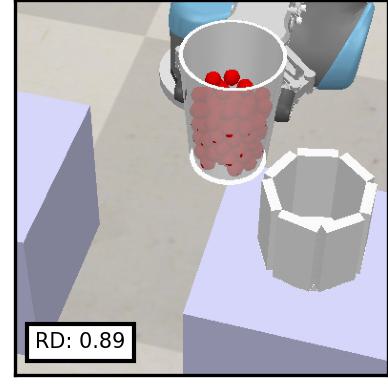}
			\caption{Alternative $RD$}
			\label{fig:alt_dd_0p2_e1}
		\end{subfigure}
		\begin{subfigure}[t]{0.24\textwidth}
			\centering
			\includegraphics[width=\linewidth]{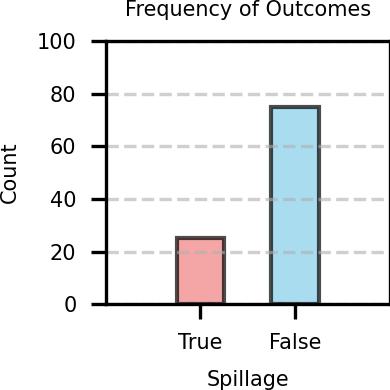}
			\caption{Outcomes with alternative $RD$}
			\label{fig:repli_alt_dd_0p2_e1}
		\end{subfigure}

		\caption{Actual causation analysis of example 1. \textbf{(a)}~Actual trial parameters of trial with spillage and \textbf{(b)}~outcome frequencies over 100 replications. \textbf{(c)}~Probabilities of actual causation inequality for $RD$. \textbf{(d)}~Probabilities of actual causation inequality for $FU$. \textbf{(e)}~Alternative $RD$ for 0.5 spillage probability and \textbf{(f)}~outcome frequencies over 100 replications. \textbf{(g)}~Alternative $RD$ for 0.2 spillage probability and \textbf{(h)}~outcome frequencies over 100 replications.}	\label{fig:ac_example_1}
	\end{figure}

	In this example, the source container was filled to a medium level ($FU=0.51$). The particles were poured into a target container of smaller capacity ($RC=0.70$) and smaller diameter ($RD=0.70$). The actual trial parameters are shown in Figure~\ref{fig:actual_e1}. The outcomes over 100 replications with the actual parameters, shown in Figure~\ref{fig:repli_actual_e1}, indicate a high probability of spillage. 
	
	Figure~\ref{fig:ac_dd_e1} shows the reference probability and the probabilities obtained for contrastive $RD$ values obtained from inequality~(\ref{eq:dd_actual_cause_inequality}). In the area where probability raising holds, the contrastive probabilities show a sharp decrease around $RD\approx0.8$ and a low probability for $RD>0.9$. We select the alternative values $RD=0.87$ (chance-level probability) and $RD=0.89$ (low probability~$\approx 0.2$), as shown in Figures \ref{fig:alt_dd_0p5_e1} and \ref{fig:alt_dd_0p2_e1}, respectively. It is important to note that changing $RV$ and keeping $RC$ constant produce a target container of smaller height. The results from 100 replications are shown in Figure~\ref{fig:repli_alt_dd_0p5_e1} and Figure~\ref{fig:repli_alt_dd_0p2_e1}. The replications with the alternative parameters show that $RD=0.89$ is better suited to avoid spillage. It is also interesting to note that due to the non-linearity of the probabilities, small changes in $RD$ yield a large effect on the probability of spillage.
	
	Figure~\ref{fig:ac_fu_e1} shows the reference probabilities and the probabilities obtained for contrastive FU values obtained from inequalities (\ref{eq:fu_actual_cause_inequality}) and (\ref{eq:fu_empty_z_actual_cause_inequality}). The probability curve that results from the contrastive values is nearly horizontal and very close in magnitude to the reference probabilities. The area for which probability raising holds (i.e., the shaded region) thus results from small magnitude differences barely noticeable in Figure~\ref{fig:ac_fu_e1}. The contrastive values for which probability raising holds yield a large probability of spillage. Therefore, selecting an alternative $FU$ is unlikely to change the outcome.

	\subsubsection{Example 2}
\begin{figure}[h]
		\centering
		\begin{subfigure}[t]{0.24\textwidth}
			\centering
			\includegraphics[width=\linewidth]{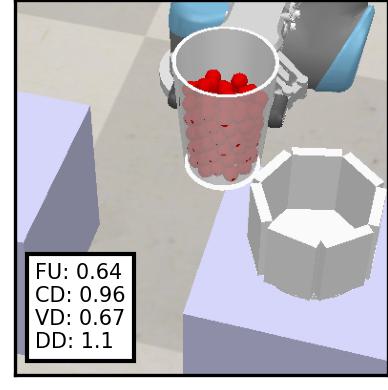}
			\caption{Actual trial parameters}
			\label{fig:actual_e2}
		\end{subfigure}  
		\begin{subfigure}[t]{0.24\textwidth}
			\includegraphics[width=\linewidth]{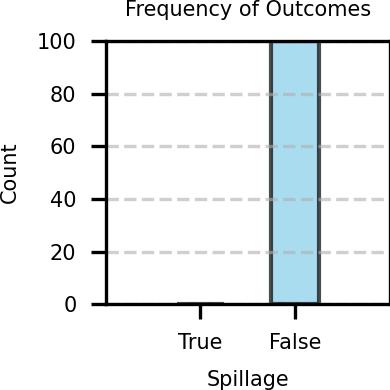}
			\caption{Outcomes with actual parameters}
			\label{fig:repli_actual_e2}
		\end{subfigure}
		\begin{subfigure}[t]{0.24\textwidth}
			\includegraphics[width=\linewidth]{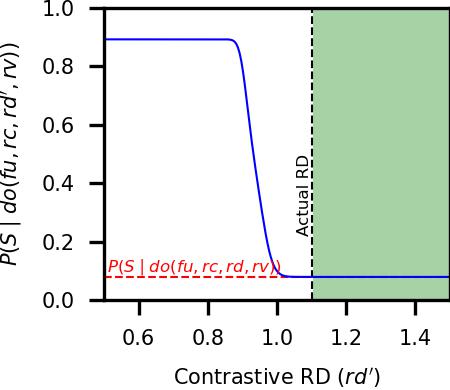}
			\caption{$RD$ as actual cause}
			\label{fig:ac_dd_e2}
		\end{subfigure}
		\begin{subfigure}[t]{0.24\textwidth}
			\includegraphics[width=\linewidth]{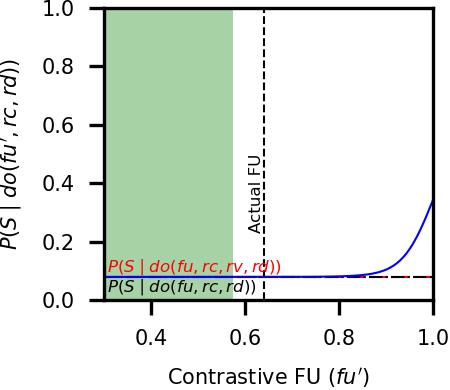}
			\caption{$FU$ as actual cause}
			\label{fig:ac_fu_e2}
		\end{subfigure}

		\caption{Actual causation analysis of example 2. \textbf{(a)}~Actual trial parameters of trial with spillage and \textbf{(b)}~outcome frequencies over 100 replications. \textbf{(c)}~Probabilities of actual causation inequality for $RD$. \textbf{(d)}~Probabilities of actual causation inequality for $FU$.}	\label{fig:ac_example_2}
	\end{figure}

	In this example, the source container was filled to a medium-high level ($FU=0.64$). The particles were poured into a target container of a slightly smaller capacity ($RC=0.96$) and slightly larger diameter ($RD=1.10$). The actual trial parameters are shown in Figure~\ref{fig:actual_e2}. The outcomes over 100 replications with the actual parameters, shown in Figure~\ref{fig:repli_actual_e2}, indicate a low probability of spillage. 
	
	Figure~\ref{fig:ac_dd_e2} shows the reference probability and the probabilities obtained for contrastive RD values obtained from inequality~(\ref{eq:dd_actual_cause_inequality}). It can be observed that the contrastive values for which probability raising holds yield a low probability of spillage. Therefore, selecting an alternative $RD$ is unlikely to change the outcome. A similar result is observed from the actual causation analysis of FU. Figure~\ref{fig:ac_fu_e2} indicates that selecting an alternative FU will unlikely change the outcome. In this example, the analysis of actual causation indicates that repeating the execution of the pouring action with the actual parameters is likely to succeed.

	\subsubsection{Example 3}
	
	In this example, the source container was filled to a high level ($FU=0.77$). The particles were poured into a target container of smaller capacity ($RC=0.79$) and slightly smaller diameter ($RD=0.98$). The actual trial parameters are shown in Figure~\ref{fig:actual_e3}. The outcomes over 100 replications with the actual parameters, shown in Figure~\ref{fig:repli_actual_e3}, indicate a moderately larger probability of spillage ($\approx 0.6$). 
	
	Figure~\ref{fig:ac_dd_e3} shows the reference probability and the probabilities obtained for contrastive $RD$ values obtained from inequality~(\ref{eq:dd_actual_cause_inequality}). In the area where probability raising holds, the contrastive probabilities show a sharp decrease around $RD\approx1.0$ and a low probability for $RD>1.1$. We select the alternative values $RD=0.99$ (chance-level probability) and $RD=1.02$ (low probability $\approx 0.2$), as shown in Figures \ref{fig:alt_dd_0p5_e3} and \ref{fig:alt_dd_0p2_e3}, respectively. The results from 100 replications are shown in Figure~\ref{fig:repli_alt_dd_0p5_e3} and Figure~\ref{fig:repli_alt_dd_0p2_e3}. The replications with the alternative parameters show that $RD=1.02$ is better suited to avoid spillage.

	Figure~\ref{fig:ac_fu_e3} shows the reference probabilities and the probabilities obtained for contrastive FU values obtained from inequalities (\ref{eq:fu_actual_cause_inequality}) and (\ref{eq:fu_empty_z_actual_cause_inequality}). In the area where probability raising holds, we observe low probabilities for $FU<0.6$. For $FU>0.6$, we observe smooth probability increments. We select the alternative values $FU=0.75$ (chance-level probability) and $FU=0.64$ (low probability $\approx 0.2$), as shown in Figures \ref{fig:alt_fu_0p5_e3} and \ref{fig:alt_fu_0p2_e3}, respectively. The results from 100 replications are shown in Figure~\ref{fig:repli_alt_fu_0p5_e3} and Figure~\ref{fig:repli_alt_fu_0p2_e3}. The replications with the alternative parameters show that $FU=0.64$ is better suited to avoid spillage.
	
	\begin{figure}[h]
		\centering
		\begin{subfigure}[t]{0.24\textwidth}
			\centering
			\includegraphics[width=\linewidth]{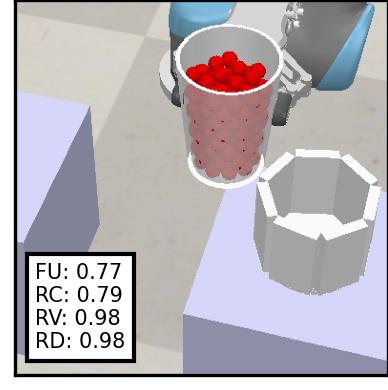}
			\caption{Actual trial parameters}
			\label{fig:actual_e3}
		\end{subfigure}  
		\begin{subfigure}[t]{0.24\textwidth}
			\includegraphics[width=\linewidth]{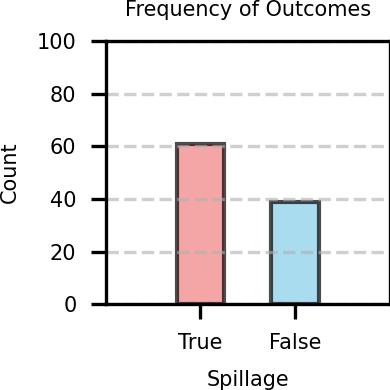}
			\caption{Outcomes with actual parameters}
			\label{fig:repli_actual_e3}
		\end{subfigure}
		\begin{subfigure}[t]{0.24\textwidth}
			\includegraphics[width=\linewidth]{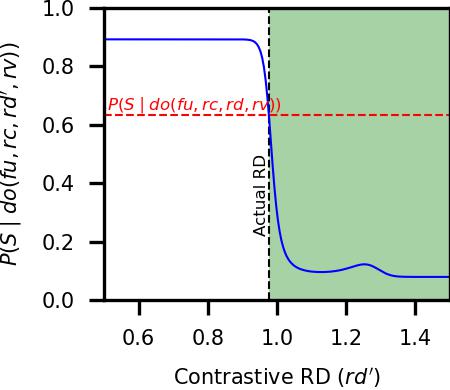}
			\caption{$RD$ as actual cause}
			\label{fig:ac_dd_e3}
		\end{subfigure}
		\begin{subfigure}[t]{0.24\textwidth}
			\includegraphics[width=\linewidth]{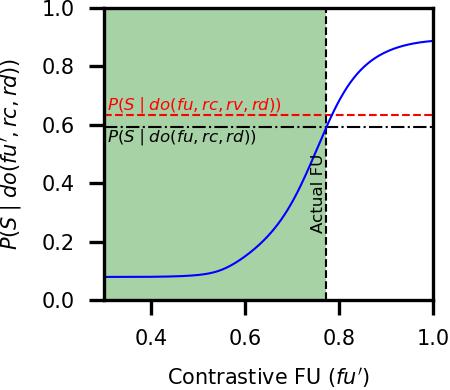}
			\caption{$FU$ as actual cause}
			\label{fig:ac_fu_e3}
		\end{subfigure}
		\\\vspace{0.5cm}
		\begin{subfigure}[t]{0.24\textwidth}
			\centering
			\includegraphics[width=\linewidth]{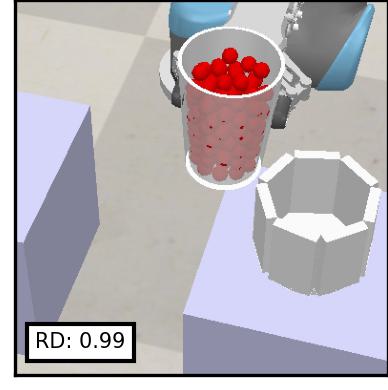}
			\caption{Alternative $RD$ }
			\label{fig:alt_dd_0p5_e3}
		\end{subfigure}
		\begin{subfigure}[t]{0.24\textwidth}
			\centering
			\includegraphics[width=\linewidth]{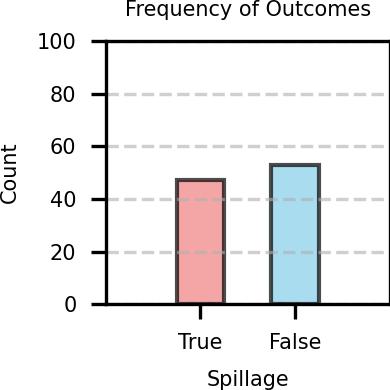}
			\caption{Outcomes with alternative $RD$}
			\label{fig:repli_alt_dd_0p5_e3}
		\end{subfigure}
		\begin{subfigure}[t]{0.24\textwidth}
			\centering
			\includegraphics[width=\linewidth]{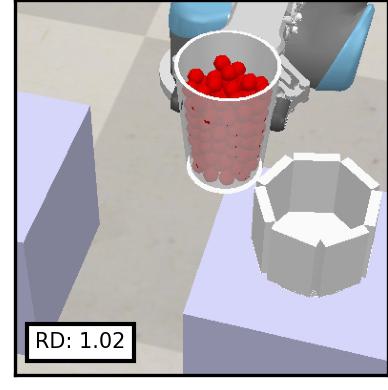}
			\caption{Alternative $RD$ }
			\label{fig:alt_dd_0p2_e3}
		\end{subfigure}
		\begin{subfigure}[t]{0.24\textwidth}
			\centering
			\includegraphics[width=\linewidth]{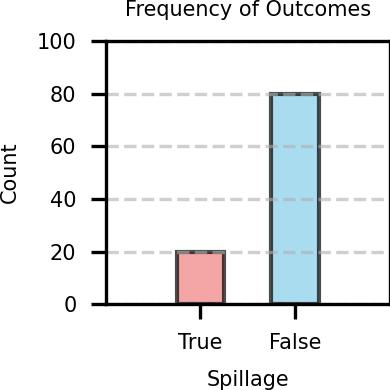}
			\caption{Outcomes with alternative $RD$}
			\label{fig:repli_alt_dd_0p2_e3}
		\end{subfigure}
		\\\vspace{0.5cm}
		\begin{subfigure}[t]{0.24\textwidth}
			\centering
			\includegraphics[width=\linewidth]{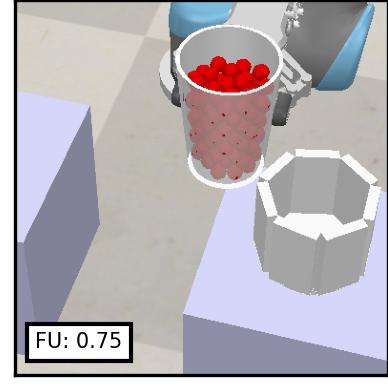}
			\caption{Alternative $FU$}
			\label{fig:alt_fu_0p5_e3}
		\end{subfigure}
		\begin{subfigure}[t]{0.24\textwidth}
			\centering
			\includegraphics[width=\linewidth]{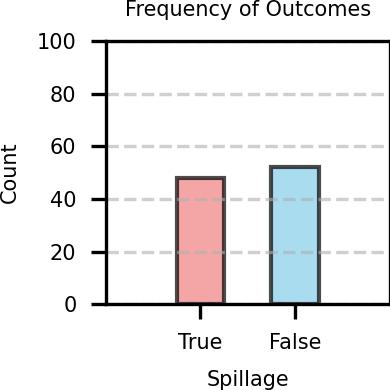}
			\caption{Outcomes with alternative $FU$}
			\label{fig:repli_alt_fu_0p5_e3}
		\end{subfigure}
		\begin{subfigure}[t]{0.24\textwidth}
			\centering
			\includegraphics[width=\linewidth]{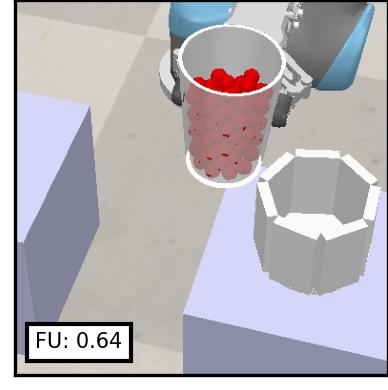}
			\caption{Alternative $FU$}
			\label{fig:alt_fu_0p2_e3}
		\end{subfigure}
		\begin{subfigure}[t]{0.24\textwidth}
			\centering
			\includegraphics[width=\linewidth]{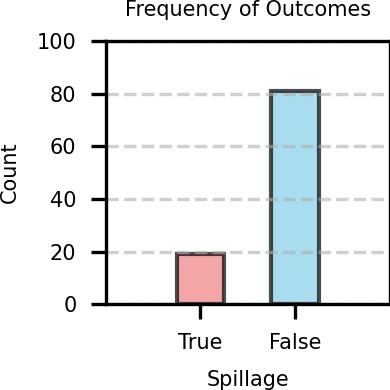}
			\caption{Outcomes with alternative $FU$}
			\label{fig:repli_alt_fu_0p2_e3}
		\end{subfigure}

		\caption{Actual causation analysis of example 3. \textbf{(a)}~Actual trial parameters of trial with spillage and \textbf{(b)}~outcome frequencies obtained over 100 replications. \textbf{(c)}~Probabilities of actual causation inequality for $RD$. \textbf{(d)}~Probabilities of actual causation inequality for $FU$. \textbf{(e)}~Alternative $RD$ for 0.5 spillage probability and \textbf{(f)}~outcome frequencies over 100 replications. \textbf{(g)}~Alternative $RD$ for 0.2 spillage probability and \textbf{(h)}~outcome frequencies over 100 replications. \textbf{(i)}~Alternative $FU$ for 0.5 spillage probability and \textbf{(j)}~outcome frequencies over 100 replications. \textbf{(k)}~Alternative $FU$ for 0.2 spillage probability and \textbf{(l)}~outcome frequencies over 100 replications.}	\label{fig:ac_example_3}
	\end{figure}
	
	\subsubsection{Example 4}
	
	In this example, the source container was filled to a high level ($FU=0.91$). The particles were poured into a target container of smaller capacity ($RC=0.69$) and slightly larger diameter ($RD=1.08$). The actual trial parameters are shown in Figure~\ref{fig:actual_e4}. The outcomes over 100 replications with the actual parameters, shown in Figure~\ref{fig:repli_actual_e4}, indicate a large probability of spillage. 
	
	Figure~\ref{fig:ac_dd_e4} shows the reference probability and the probabilities obtained for contrastive $RD$ values obtained from inequality~(\ref{eq:dd_actual_cause_inequality}). It can be observed that the contrastive values for which probability raising holds yield a high probability of spillage. Therefore, selecting an alternative $RD$ is unlikely to change the outcome.

	Figure~\ref{fig:ac_fu_e4} shows the reference probabilities and the probabilities obtained for contrastive FU values obtained from inequalities (\ref{eq:fu_actual_cause_inequality}) and (\ref{eq:fu_empty_z_actual_cause_inequality}). In the area where probability raising holds, we observe low probabilities for $FU<0.4$. For $FU>0.4$, we observe smooth probability increments. We select the alternative values $FU=0.76$ (chance-level probability) and $FU=0.53$ (low probability $\approx 0.2$), as shown in Figures \ref{fig:alt_fu_0p5_e4} and \ref{fig:alt_fu_0p2_e4}, respectively. The results from 100 replications are shown in Figure~\ref{fig:repli_alt_fu_0p5_e4} and Figure~\ref{fig:repli_alt_fu_0p2_e4}. The replications with the alternative parameters show that $FU=0.53$ is better suited to avoid spillage.

	\begin{figure}[h]
		\centering
		\begin{subfigure}[t]{0.24\textwidth}
			\centering
			\includegraphics[width=\linewidth]{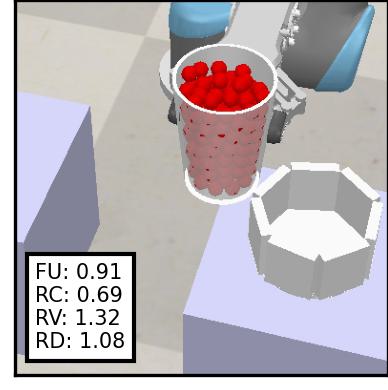}
			\caption{Actual trial parameters}
			\label{fig:actual_e4}
		\end{subfigure}  
		\begin{subfigure}[t]{0.24\textwidth}
			\includegraphics[width=\linewidth]{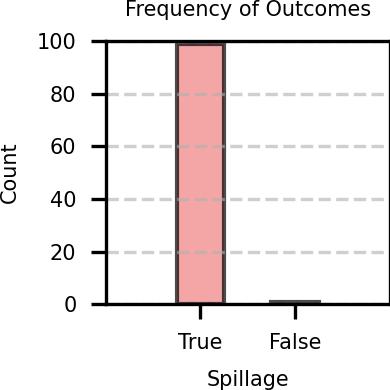}
			\caption{Outcomes with actual parameters}
			\label{fig:repli_actual_e4}
		\end{subfigure}
		\begin{subfigure}[t]{0.24\textwidth}
			\includegraphics[width=\linewidth]{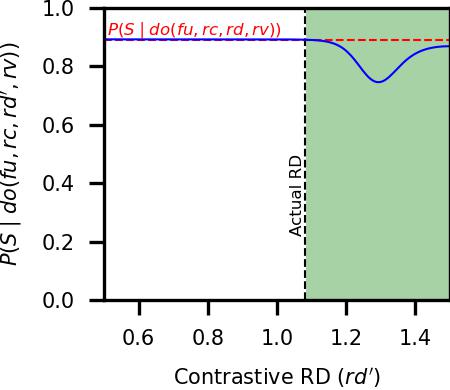}
			\caption{$RD$ as actual cause}
			\label{fig:ac_dd_e4}
		\end{subfigure}
	\begin{subfigure}[t]{0.24\textwidth}
			\includegraphics[width=\linewidth]{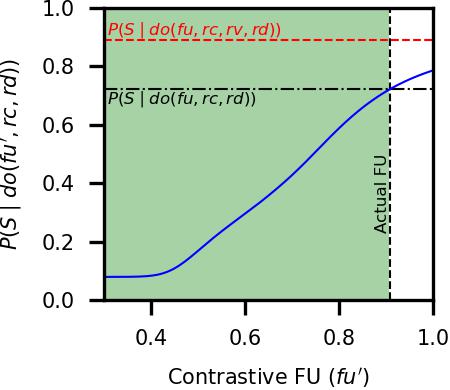}
			\caption{$FU$ as actual cause}
			\label{fig:ac_fu_e4}
		\end{subfigure}
		\\\vspace{0.5cm}	
		\begin{subfigure}[t]{0.24\textwidth}
			\centering
			\includegraphics[width=\linewidth]{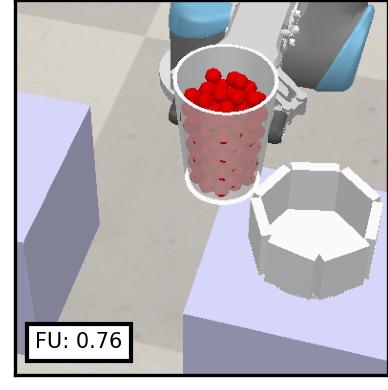}
			\caption{Alternative $FU$}
			\label{fig:alt_fu_0p5_e4}
		\end{subfigure}
		\begin{subfigure}[t]{0.24\textwidth}
			\centering
			\includegraphics[width=\linewidth]{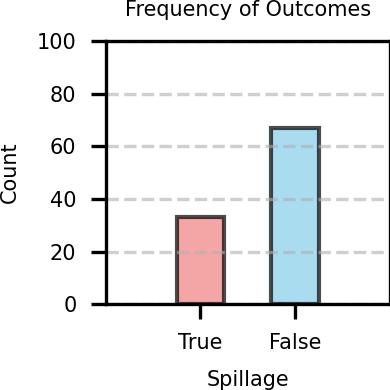}
			\caption{Outcomes with alternative $FU$}
			\label{fig:repli_alt_fu_0p5_e4}
		\end{subfigure}
		\begin{subfigure}[t]{0.24\textwidth}
			\centering
			\includegraphics[width=\linewidth]{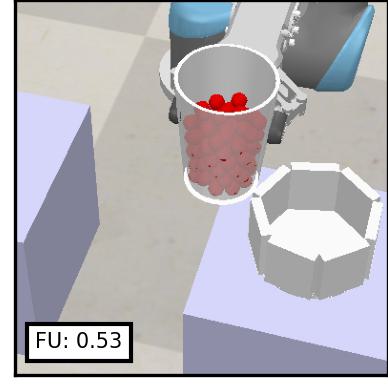}
			\caption{Alternative $FU$}
			\label{fig:alt_fu_0p2_e4}
		\end{subfigure}
		\begin{subfigure}[t]{0.24\textwidth}
			\centering
			\includegraphics[width=\linewidth]{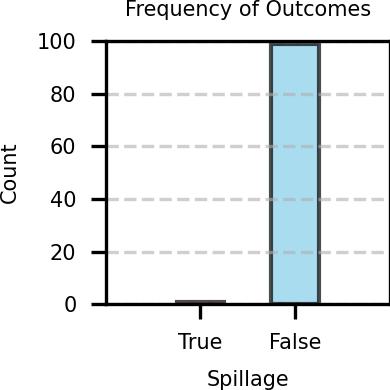}
			\caption{Outcomes with alternative $FU$}
			\label{fig:repli_alt_fu_0p2_e4}
		\end{subfigure}
		
		\setcounter{figure}{9}
		\setcounter{subfigure}{-1}
		\caption{Actual causation analysis of example 4. \textbf{(a)}~Actual trial parameters of trial with spillage and \textbf{(b)}~outcome frequencies over 100 replications. \textbf{(c)}~Probabilities of actual causation inequality for $RD$. \textbf{(d)}~Probabilities of actual causation inequality for $FU$. \textbf{(e)}~Alternative $FU$ for 0.5 spillage probability and \textbf{(f)}~outcome frequencies over 100 replications. \textbf{(g)}~Alternative $FU$ for 0.2 spillage probability and \textbf{(h)}~outcome frequencies over 100 replications.}
		\label{fig:ac_example_4}
	\end{figure}
	
	\subsection{Evaluation of alternative actions to prevent spillage}\label{sec:evaluation_corrective_actions}

	In this section, we evaluate the capabilities of the analysis of probabilistic actual causation to guide the selection of alternative parameters to prevent spillage. For this evaluation, we generated a test dataset of 3000 pouring trials. The trial parameters of the test dataset were sampled from the same distributions used for the training dataset, as described in Table~\ref{table:variables}. Spillage occurred in 1216 trials and 1784 trials were successful.

	Following steps described in Section~\ref{sec:how_to_use_actual_causality}, we conducted the analysis of probabilistic actual causation on the spillage trials. We identified the range of contrastive values where probability raising holds, and within these values, we selected the subset of values with probability of spillage$<0.1$. From this subset, the closest value to the current parameter was selected as alternative parameter. We ran the trial using the alternative parameter and recorded the outcome.

	As explained in Section~\ref{sec:how_to_use_actual_causality}, the analysis of actual causation can be applied to different variables, one at a time. When inequality~(\ref{eq:actual_cause_inequality}) does not hold for any of the contrastive values, the variable cannot be regarded as an actual cause. This indicates that there are no alternative values for the analyzed variable. Even if inequality~(\ref{eq:actual_cause_inequality}) holds, it can also occur that the probabilities within the range of contrastive values where probability raising holds lay above the chance level or above the desired probability threshold (0.1 in our case). Therefore, it may happen that no alternative values for the variable being analyzed can be identified. To evaluate this aspect, we conducted the analysis of probabilistic actual causation on the $RC$, $FU$, and $RD$ variables, and determined the percentage of the spillage trials for which an alternative parameter satisfying the probability threshold could be identified. For $RC$, an alternative parameter could be identified in 2.7\% of the spillage trials, for $FU$ in 59\% of the spillage trials, and for $RD$ in 97.9\% of the spillage trials.

	The marked differences between variables can be attributed to the causal structure of the task and the magnitude of the causal probabilities presented in Section~\ref{sec:causal_effect_results}. $RC$ has an indirect effect on $S$ through $RV$. $FU$ also has an indirect effect on $S$ through $RV$ and a direct effect. Comparing the causal probabilities $P(S|do(RC))$ (Figure~\ref{fig:doCD}), with $P(S|do(FU))$ and $P(S|do(FU, RV ))$ (Figures~\ref{fig:doFU} and~\ref{fig:doFUdirect}, respectively), it can be observed that the effect of $FU$ on $S$ is stronger than the effect of $RC$. Under these considerations, finding an alternative $RC$ value is less likely than finding an alternative $FU$ value. In contrast to $RC$ and $FU$, $RD$ has a direct effect on $S$, and the $P(S|do(RD))$ (Figure~\ref{fig:doDD}) shows a range with low spillage probability, which leads to a higher likelihood of finding an alternative value.

	Next, we evaluate the pouring success rates obtained by running the spillage trials using the alternative $RD$ and $FU$ values, while keeping the other variables unchanged \footnote{$RC$ was not considered for this evaluation due to the low number of trials for which an alternative value was found.}. As explained at the beginning of this section, in each spillage trial we set $RD$ or $FU$ to an alternative value (closest to the current value) predicted to have a spillage probability $<0.1$ according to the actual causation analysis. A trial was successful if all the particles were poured into the target container. From 1216 spillage trials, an alternative $RD$ value was identified for 1191 trials. Running these trials with the alternative DD values produced a success rate of 88.7\%. From 1216 spillage trials, an alternative $FU$ value was identified for 718 trials. Running these trials with the alternative $FU$ values success rate of  86.9\%. The success rates demonstrate the practical value of the actual causation approach in identifying alternative parameters.

\begin{figure}[h]	
		\centering
	\begin{subfigure}[b]{0.45\textwidth}
			\includegraphics[width=\linewidth]{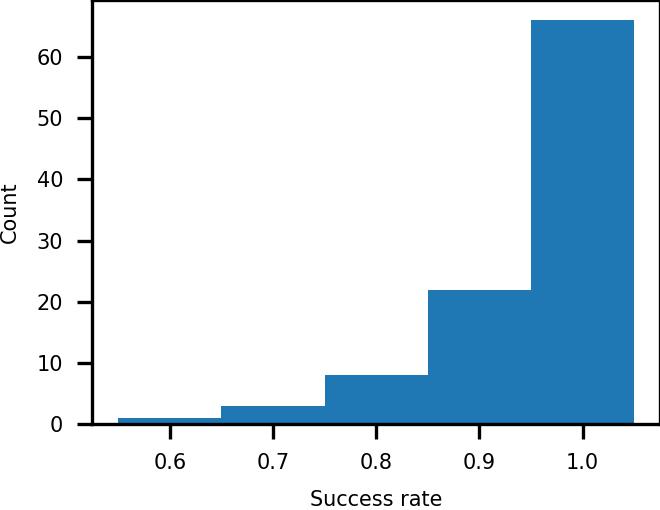}
			\caption{Empirical success rates using alternative $RD$ values.}
			\label{fig:dd_success_rate_hist}
	\end{subfigure}  	
		\begin{subfigure}[b]{0.45\textwidth}
			\includegraphics[width=\linewidth]{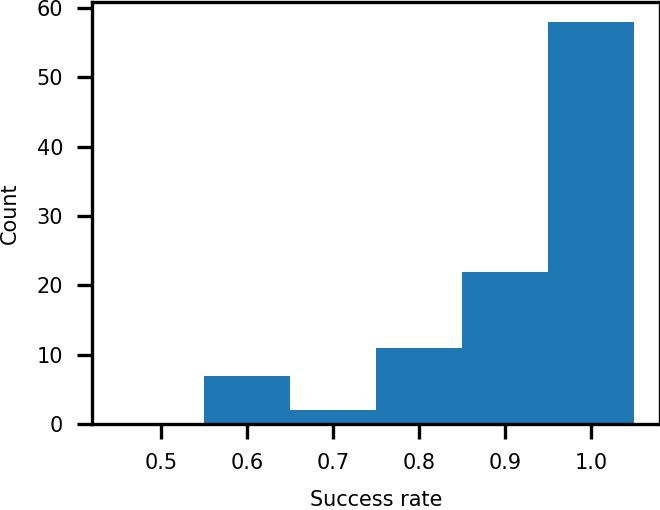}
			\caption{Empirical success rates using alternative $FU$ values.}
			\label{fig:fu_success_rate_hist}
		\end{subfigure}
		\caption{Empirical success rates obtained over 10 replications. }	\label{fig:success_rates_histograms}
	\end{figure}

	Finally, we evaluate the empirical success rates observed when conducting trial replications using the alternative $FU$ and $RD$ values. For this, we selected a random subset of 100 spillage trials from the test dataset. We ran 10 replications of each trial of the subset using the alternative $RD$ or $FU$ values. As a result, we obtained an empirical success rate for each trial ($ \frac{number \ of \ successful \ replications}{10}$). The histogram of the empirical success rates obtained with the alternative $RD$ values is shown in Figure~\ref{fig:dd_success_rate_hist}, and the histogram obtained for the alternative $FU$ values is shown in Figure~\ref{fig:fu_success_rate_hist}. In general, the observed success rates are in line with the probability criterion used to select the alternative values. Nevertheless, lower success rates of 0.6 and 0.7 were also obtained for some trials. Crucially, all the success rates are above the chance level, which provides empirical support for the usefulness of the alternative values.
	
	\section{Discussion}\label{sec:discussion}
	In Section~\ref{sec:actual_causality_results}, an analysis of actual causation was conducted to determine the actual cause of spillage in a set of selected spillage trials. The analysis of the $FU$ and $RD$ variables was conducted individually. Based on the analysis, alternative $FU$ and $RD$ values were selected. In the examples, we also observed cases in which the analysis indicated that no alternative $FU$ or $RD$ values would significantly reduce the probability of spillage. 
	
	The actual causation probabilities exhibit a non-linear behavior. In particular, a sharp decrease in the spillage probability can be observed in some ranges of $RD$. Considering the sharpness in the transition from low to high probability of spillage observed in the causal probability $P(S|do(RD))$ (Figure~\ref{fig:doDD}) and in the actual causation probabilities in examples 1, 2, and 3 (Figures \ref{fig:ac_dd_e1}, \ref{fig:ac_dd_e2}, and \ref{fig:ac_dd_e3}, respectively), we verified that the sharp transition is not an artifact of the binary representation of the outcome. For this verification, we examined the number of particles spilled and the percentage of spilled particles as a function of $RD$ in the range with sharp probability transitions. Within this range of values, small changes in $RD$ yield significant changes in the number of spilled particles and the spilled percentage. As a result of the sharp probability decrease, when the analysis indicates that a $RD$ is the actual cause of spillage, small changes in its value significantly impact the probability of spillage (cf. examples 1 and 3). The actual causation probabilities of $FU$ also exhibit non-linear behavior, though the ranges where the probability of spillage transitions from low to high show a less pronounced slope (cf. example 4). Nevertheless, there are cases where a small change in $FU$ leads to significant differences in the probability of spillage (cf. example 3).
	
\begin{figure}[h!]
		\begin{center}
			\includegraphics[width=1\textwidth]{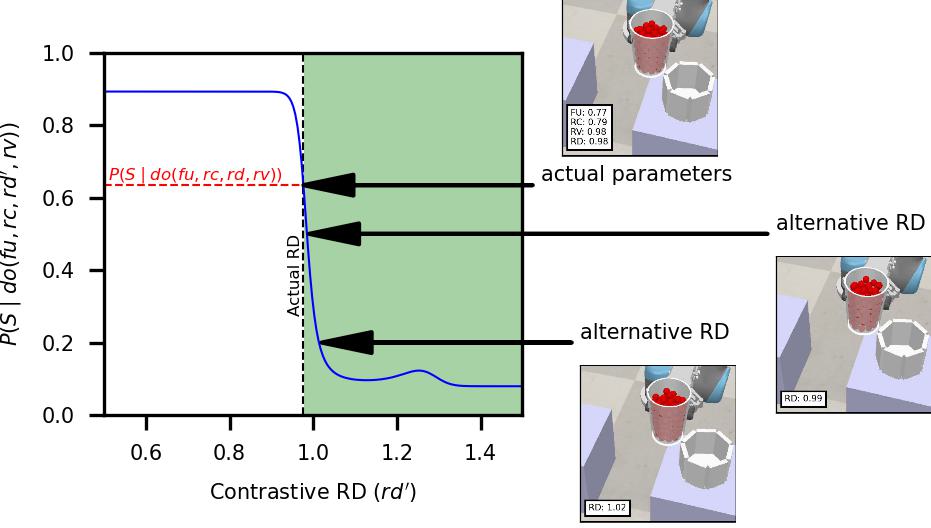}
		\end{center}
		\caption{Actual causation probabilities, actual trial parameters and target containers with different $RD$ values.}\label{fig:discussion_DD_comparison}
	\end{figure}

	The examples show that the probabilities of actual causation provide a principled criterion for comparing alternative actions with respect to the probability of the desired outcome. In the pouring task, small differences between values separate a ``bad" from a ``good" alternative action due to the non-linear effect of the variables on the outcome. Based on this consideration, it is reasonable to assume that the alternative actions identified using the automatic application of the actual causation analysis might differ from those chosen by a human observer. Consider the target container dimensions from example 3 compared in Figure~\ref{fig:discussion_DD_comparison}. A human observer might fail to realize that a slight change in diameter can significantly reduce the chances of spillage due to non-linear interaction between the task variables. Therefore, it can be assumed that a human observer will choose alternative actions based on larger parameter differences than the ones suggested by the analysis of actual causation. For a human observer, a significantly larger diameter difference or a lower fullness level might result from an implicit safety margin in the selection of an alternative action to avoid spillage. Overall, human reasoning about alternative solutions will hardly resemble the analysis of actual causation, as it relies on probabilistic reasoning about the (non-linear) interaction between variables. 
	
	Regarding the usage of actual causation for action guidance, it is important to consider the availability or feasibility of alternative actions in the context of application. In simulation, generating target containers of different dimensions or changing the source container's fullness levels is straightforward. However, the available alternative actions might be limited in the real world. For example, if target containers are available only in two diameter sizes, the selection of an alternative action is reduced to making a forced choice, leaving aside any reasoning about the effect of the diameters in a continuous space on the probability of spillage. Nevertheless, even if the alternatives are limited, an analysis of actual causation can provide the agent performing or monitoring the task with useful information about possible alternative actions.  
	
	It is important to recall that the results obtained from applying the analysis of actual causation depend on the structure of the causal graph and the variable representation. In this work, we opted to represent spillage as a binary variable. However, other representations of the outcome are possible. For example, spillage could be represented as the number of spilled marbles or as a relation, such as $S = \frac{number \ of \ spilled \ marbles}{number \ of \ marbles \ in \ the \ source \ container}$. Defining spillage as a binary variable treats spilling one or many marbles equally. In this sense, the binary representation loses information regarding the severity of spillage. Ideally, the perfect realization of the task entails pouring without spillage. However, whether information about the severity of spillage is necessary depends on the context. For example, while spilling a few snacks at a party would not be a problem, spilling a single particle in a chemistry laboratory might be inadmissible. The previous situational examples emphasize that the context of the application must be considered when determining the definition of the outcome variable(s).
	
	The representation of the variables has implications for the perceptual capabilities of the agent performing or monitoring the task, be it a human or a robot. The perception of the outcome and the container properties relies on sensory cues, which might consist of visual and force feedback. For example, to determine the number of spilled particles, the agent must be able to perceive and count individual particles. The extent to which this is feasible depends on the context (e.g., counting spilled candies might be way easier than counting spilled rice grains, both for a human or a robot). In this respect, representing spillage as a binary variable has the advantage of being easier to determine, both for a human and a robot, as it requires less perception, reasoning, and action capabilities. Overall, the successful implementation of action guidance based on the analysis of probabilistic actual causation in a real-world application relies on the availability of the information necessary to compute the probabilities in inequality~(\ref{eq:actual_cause_inequality}). For the pouring task, the agent would need an accurate perception of both containers' dimensions, the source container's fullness level, and whether or not spillage occurred.  
	
	\section{Conclusions}
	
	In this paper, we conducted a probabilistic actual causation analysis of a robot pouring task. The modeling based on causal graphs and the estimation of conditional probability distributions using neural networks facilitated a qualitative and quantitative understanding of the influence of various factors on the task's outcome. Throughout a series of examples, we demonstrated that the analysis of actual causation provides a principled approach to check whether a variable is a cause of the outcome and to select alternative actions appropriate to change the observed outcome. Our results show that the analysis of actual causation provides information about the extent to which a variable caused an observed outcome that cannot be retrieved directly from simple causal probabilities. This occurs because the causal probability of a variable on the outcome lacks information about the context of the other variables. In contrast, the definition of probabilistic actual causation considers the context of the variables and their role in the causal structure  (i.e., whether the variables are mediators or are outside the causal path). In the pouring task, the analysis of actual causation enabled us to determine whether a variable was a cause of spillage and, based on the assessment of the probabilities of actual causation, the selection of an alternative action parameter. 
	
	The reliability of the analysis of actual causation relies on the correctness of the causal graph structure and the estimated conditional probability distributions. This constitutes a major challenge for implementing an actual causation analysis in real-life tasks as it requires 1) a careful selection of the variables' representation, 2) determining the structure of the causal graph, and 3) estimating conditional probability distributions. The methods described in Section \ref{sec:materials_and_methods} constitute state-of-the art best common practices to obtain reliable and robust causal modeling results. Specifically, we used a realistic simulation of the pouring task to cover an ample combinatorial space of task parameters, which would have been cumbersome to replicate in a real environment. The simulation provided us with a large dataset to learn the causal structure of the task using a causal discovery algorithm with bootstrapping and to estimate its causal probability distributions using neural networks. In addition to the information provided in Section \ref{sec:materials_and_methods}, in the supplementary material we discuss further modeling assumptions and we provide empirical support to the correctness of the causal model.
	
	We demonstrated the practical use of probabilistic actual causation in a robotic task. In addition, the information retrieved from actual causation analysis can be used in the context of human-machine interaction to support human decision-making. Recalling that the actual causation probabilities can be interpreted as the extent to which an alternative action parameter is a "good" or "bad" corrective action, the framework can provide the human operator with additional information and contextual cues, for example, in augmented or virtual reality applications, to support the selection of action parameters. 
	
	In an additional scope of application, the analysis of actual causation can provide an objective baseline to evaluate the human perception of actual causes and the selection of alternative actions when a different outcome is sought. For example, the extent to which the causes perceived by a human observer correspond to the actual causes identified by the probabilistic framework can be investigated. Furthermore, the framework can be used to assess the extent to which human ratings of corrective actions in a continuum from ``bad" to ``good" correspond to the interpretation of the goodness of an alternative action parameter based on the actual causation probabilities used in this paper.  
	
	\section*{Conflict of Interest Statement}
	The authors declare that the research was conducted in the absence of any commercial or financial relationships that could be construed as a potential conflict of interest.
	
	\section*{Author Contributions}
	JM: Conceptualization, Data curation, Formal analysis, Investigation, Methodology, Software, Visualization, Writing – original draft, Writing – review \& editing; JK: Data curation, Methodology, Software, Writing – original draft; CZ: Conceptualization, Funding acquisition, Methodology, Writing – original draft, Writing – review \& editing ; VD: Conceptualization, Formal analysis, Funding acquisition, Writing – original draft, Writing – review \& editing; KS: Funding acquisition, Project administration, Supervision, Writing – review \& editing.
	
	\section*{Funding}
	The research reported in this paper has been supported by the German Research Foundation DFG, as part of Collaborative Research Center (Sonderforschungsbereich) 1320 Project-ID 329551904 “EASE - Everyday Activity Science and Engineering”, University of Bremen (\url{http://www.ease-crc.org/}). The research was conducted in subproject H01 “Sensorimotor and Causal Human Activity Models for Cognitive Architectures.”

	\section*{Acknowledgments}
	The implementation of the neural autoregressive density estimators was possible thanks the collaborative work and support of Konrad Gadzicki (EASE - subproject H03 “Discriminative and Generative Human Activity Models for Cognitive Architectures”).
	
	\bibliographystyle{unsrtnat}
	\bibliography{robot-pouring-arxiv-refs}  
	
	\section*{Supplementary Material}
	\subsection{Causal discovery using the PC algorithm - Tetrad setup and bootstrapping results}
\begin{figure}[htbp]
		\begin{center}
			\includegraphics[width=10cm]{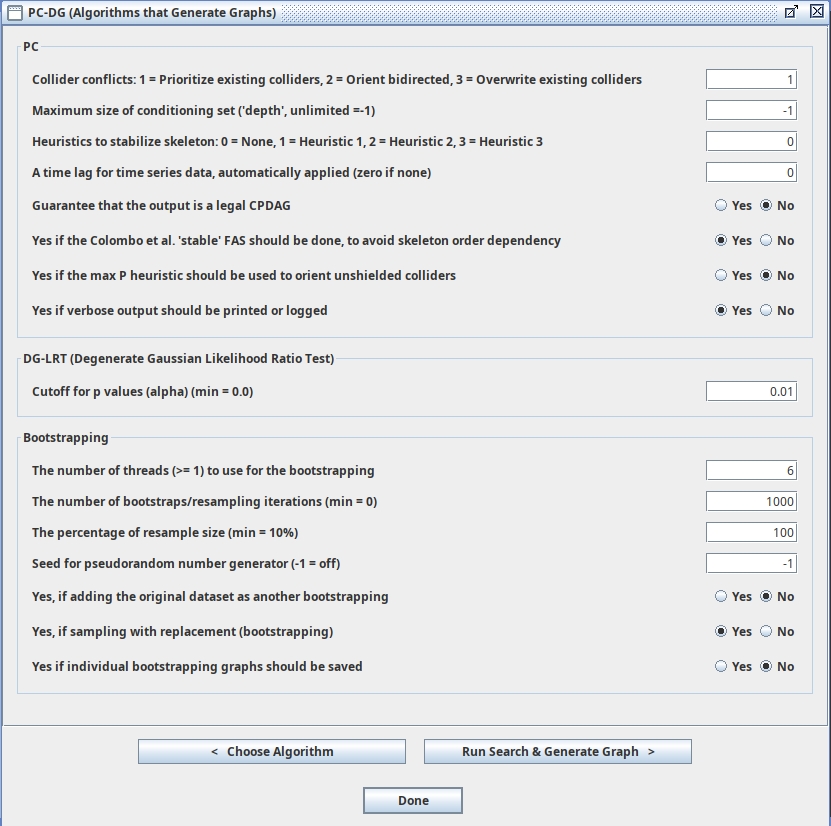}
		\end{center}
		\caption{PC and DG parameters specified in Tetrad.}\label{supfig:1}
	\end{figure}

\begin{figure}[htbp]
		\begin{center}
			\includegraphics[width=9cm]{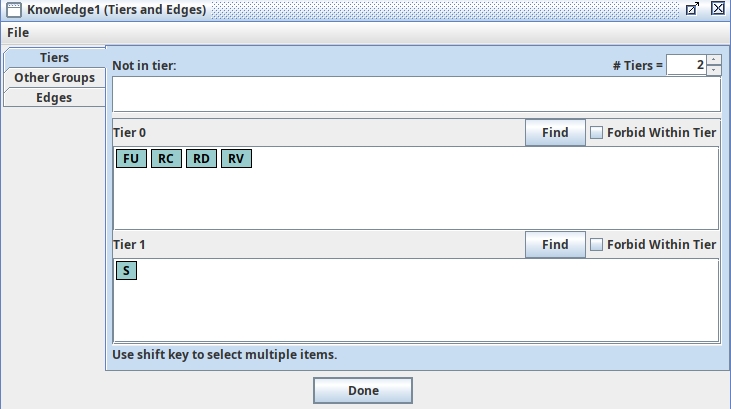}
		\end{center}
		\caption{Background knowledge specified in Tetrad.}\label{supfig:2}
	\end{figure}

	Figure \ref{supfig:1} shows the exact parameters used in Tetrad to run the PC algorithm. The specification of background knowledge in the form of tiers is shown in Figure \ref{supfig:2}. Note that within-tier edges are allowed in the background knowledge. 
	
\begin{figure}[htbp]
		\begin{center}
			\includegraphics[width=12cm]{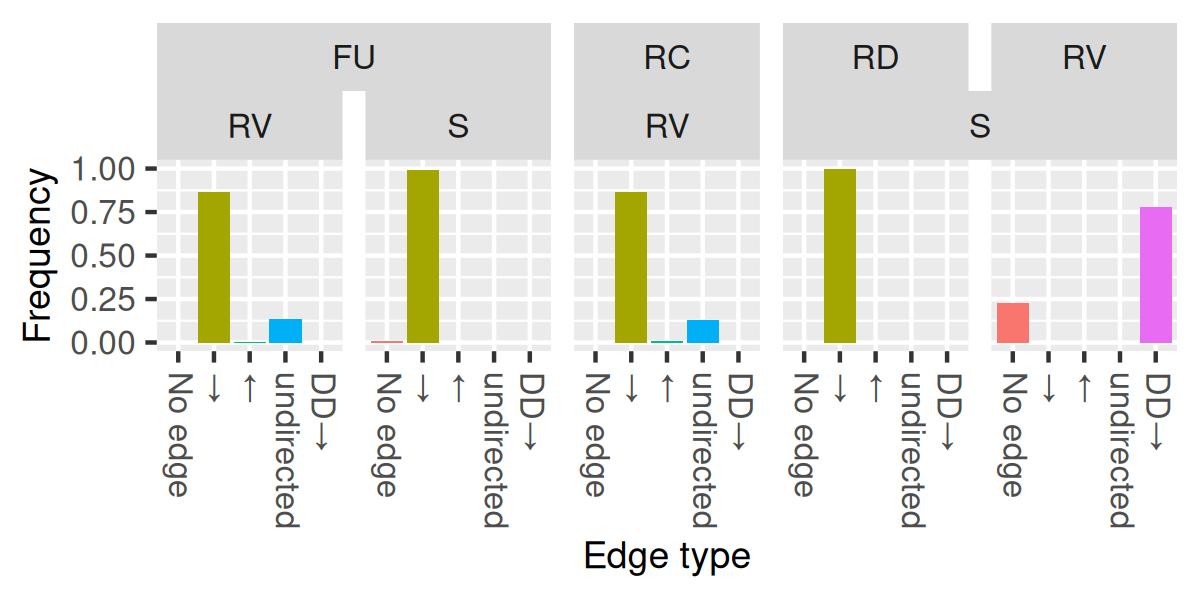}
		\end{center}
		\caption{Discovered edge type frequencies.}\label{supfig:3}
	\end{figure}

	To validate the causal relationships inferred from the data, we performed a bootstrapping analysis. Conducting a bootstrapping analysis is recommended for causal discovery\citep{MalinskyDanks2017,Glymour2019}. We ran the PC algorithm on 1000 bootstraps, producing 1000 different structures. The edge-type frequencies obtained from bootstrapping indicate whether the discovered causal relationships are stable across different bootstrap samples~\citep{Glymour2019}. In Figure \ref{supfig:3}, we report the frequency of the edge types between variables. For ease of interpretation and comparison, the frequency of edge type is reported as a proportion of the number of bootstraps. Given the different edge types (including ``no edge''), we interpret an edge frequency larger than 0.5 as stable. The reported DAG, termed \textit{discovered DAG} includes the stable edges and no-edges. The PC algorithm discovered direct edges, edges that are definitely direct (DD), and undirected edges (i.e., the data are consistent with $X \rightarrow Y$ and $X \leftarrow Y$ \citep{MalinskyDanks2017}). 
	
	The reported DAG was constructed from the edges with frequencies larger than 0.5. As shown in Figure~\ref{supfig:3}, the discovered DAG results from edge frequencies larger than 0.75. This increases our confidence in the correctness of the discovered causal structure, which is crucial for the analysis of probabilistic actual causation.

	Internally, we also conducted the causal discovery analysis with bootstrapping using Tetrad's GFCI, FGES, GRASP, and BOSS algorithm implementations (for all algorithms, we used the Degenerate Gaussian Likelihood Ratio Test (DG-LRT)). The edge frequencies slightly vary across algorithms, but the discovered structure is the same. For simplicity, we opted to report the PC results since it is a well-established algorithm with extensively studied properties~\citep{MalinskyDanks2017,Glymour2019}.
	
	\subsection{Assumption of no latent variables}

	The PC algorithm assumes no latent confounders or unobserved variables. In this respect, we rely on our analysis of the task and the data-generating process to support this assumption.

	Based on the analysis of the task described in Section ``4.2 DAG Variables" we assume that the variables used to represent the data-generating process of the simulation capture all the relevant causes of spillage. The variables represent the stochastic effect of 1) the containers' characteristics (capacities and diameters, expressed as the variables $RC$ and $RD$) and 2) the poured amount (fullness and volume, expressed as the variables $FU$ and $RV$) on the probability of spilling (variable $S$). The randomness of the outcome $S$ results from the interplay between $RC$, $RD$, $FU$, and $RV$ and the behavior of the particles during the pouring movement. The parameters the physics engine uses (particle size and density) to simulate the particles' behavior have a constant value. Therefore, they are not included as DAG variables.

	Additionally, the pouring movement was executed with constant rotation velocity and angle. Thus, these were not included as DAG variables. Based on these considerations, we are confident that we included all the relevant random variables of the data-generating process of the simulation and assume that there are no latent confounders or other unobserved variables with a causal effect on spillage.

	\subsection{Implementation of the neural autoregressive density estimators}
	
	The implementation of the NADE networks is based on the source code provided by Garrido (2021), publicly available in \url{https://github.com/Chechgm/causal_effect_estimation_using_nade} (access: 05.12.24). The following parameters were used:
	
	\begin{itemize}
		\item Neural network architecture: 2 hidden layers with 16 units each
		\item Activation function:  hyperbolic tangent (Tanh)
		\item Optimizer: RMSProp
		\item Learning rate: 0.01
	\end{itemize}

	The neural networks were implemented using Pytorch \citep{pytorch} (Version 1.10.0). 
	
	The hyper-parameters were selected based on the results reported by \citet{Garrido2021}. In their extensive analysis of the performance of different hyper-parameters (hidden layers, number of units, and learning rates), they conclude that no single combination of hyper-parameters is superior to others. We selected two hidden layers with 16 units. We noticed that increasing the number of layers and units did not reduce the loss or improve the prediction performance of the network. Based on this empirical observation, which is in line with the observations of \citet{Garrido2021}, we abstained from conducting any further systematic search or comparison of the hyper-parameter space.
	
	\subsection{Empirical support to the correctness of the causal model}

	The analysis of probabilistic actual causation relies on the correctness of the causal structure and the estimated causal probabilities. It is important to note that the causal structure of the data-generating process is unknown. Therefore, no ground truth causal graph is available to benchmark the structure and the causal probabilities learned from the training dataset.

\begin{figure}[h]
		\begin{center}
			\includegraphics[width=7cm]{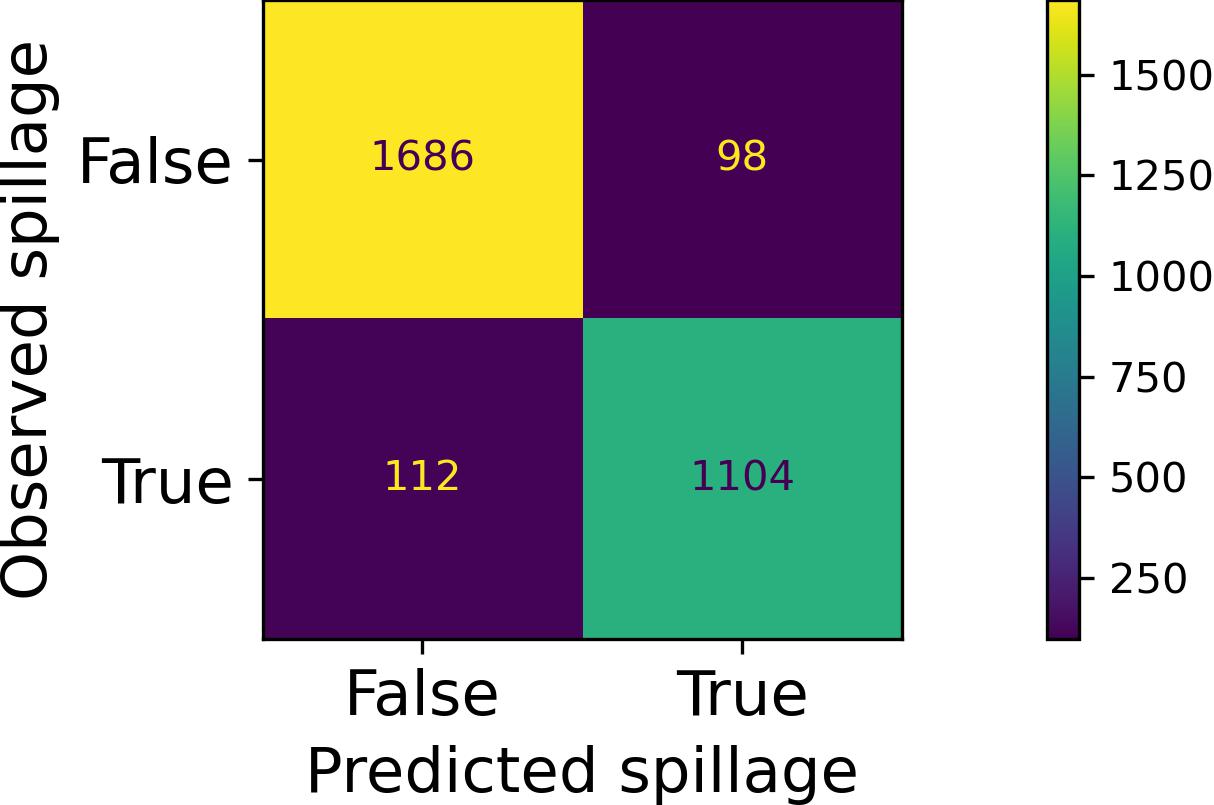}
		\end{center}
		\caption{Confusion matrix.}\label{supfig:4}
	\end{figure}

	In simulation, we have control over the parameters used in each pouring trial. The actual movement of the particles during the pouring movement, which determines whether or not spillage occurs, depends on the interaction between the trial parameters and the physics simulation. This interaction between the trial parameters and the spillage outcome is unknown. Therefore, the benchmarking of the causal graph can only be conducted on the level of evaluating the spillage predictions against the ground truth. We evaluate the spillage predictions on a test dataset of 3000 pouring trials. The trial parameters of the test dataset were sampled from the same distribution used for the training dataset (see description in Section ``4.2 DAG Variables").  For each trial, we compute the causal probability of spillage $P(S|do(FU,RC,RD,RV))$. If $P(S|do(FU,RC,RD,RV)) \geq 0.5$, the outcome prediction is labeled as $S=True$. We compare the predicted outcome with the actual outcome. The prediction results are summarized in the confusion matrix shown in Figure~\ref{supfig:4}. Among the spillage trials, the causal model predictions yield 90.8\% true positives and 9.2\% false negatives. On the other hand, the causal model predictions yield 94.5\% true negatives and 5.5\% false positives among the no-spillage trials. These results provide empirical support that the causal model corresponds with the ground truth data-generating process.

\end{document}